\documentclass[acmsmall,authorversion,nonacm]{acmart}    

\usepackage{algorithm}
\usepackage{algorithmicx}
\usepackage{footnote}
\makesavenoteenv{tabular}
\makesavenoteenv{table}
\usepackage[noend]{algpseudocode}
\usepackage{amsthm}
\usepackage{multirow}
\usepackage{pifont}
\newtheorem{definition}{Definition}[section]
\newcommand{\sstitle}[1]{\vspace{1mm} \noindent {\bf #1}}

\AtBeginDocument{%
  \providecommand\BibTeX{{%
    \normalfont B\kern-0.5em{\scshape i\kern-0.25em b}\kern-0.8em\TeX}}}
\newcommand{\cmark}{{\small\ding{51}}}
\newcommand{\xmark}{{\small\ding{55}}}

\newcommand{\dashed}{--} 

\definecolor{lightgreengray}{RGB}{200, 220, 200}
\definecolor{pastelgreen}{RGB}{173, 227, 181}
\definecolor{flashygreen}{RGB}{0, 255, 0}
\newcommand{\lightgreenpoint}{\textcolor{lightgreengray}{\normalsize$\bullet$}}
\newcommand{\pastelgreenpoint}{\textcolor{pastelgreen}{\LARGE$\bullet$}}
\newcommand{\flashygreenpoint}{\textcolor{flashygreen}{\Huge$\bullet$}}
\usepackage{subfiles}
\setcopyright{acmcopyright}
\copyrightyear{2018}
\acmYear{2018}
\acmDOI{XXXXXXX.XXXXXXX}

\begin{document}

\title[Beyond Transduction: A Survey on Inductive, Few Shot, and Zero Shot Link Prediction in KGs]{Beyond Transduction: A Survey on Inductive, Few Shot, and Zero Shot Link Prediction in Knowledge Graphs}

\author{Nicolas Hubert}
\authornote{Corresponding author}
\email{nicolas.hubert@univ-lorraine.fr}
\orcid{0000-0002-4682-422X}
\affiliation{%
  \institution{Université de Lorraine, ERPI, CNRS, LORIA}
  \city{Nancy}
  \country{France}
}

\author{Pierre Monnin}
\orcid{0000-0002-2017-8426}
\email{pierre.monnin@inria.fr}
\affiliation{%
  \institution{Université Côte d’Azur, Inria, CNRS, I3S}
  \city{Sophia Antipolis}
  \country{France}
}

\author{Heiko Paulheim}
\authornote{Corresponding author}
\email{heiko.paulheim@uni-mannheim.de}
\orcid{0000-0003-4386-8195}
\affiliation{%
  \institution{University of Mannheim}
  \city{Mannheim}
  \country{Germany}}

\renewcommand{\shortauthors}{Hubert et al.}

\begin{abstract}
Knowledge graphs (KGs) comprise entities interconnected by relations of different semantic meanings. KGs are being used in a wide range of applications. However, they inherently suffer from incompleteness, \textit{i.e.} entities or facts about entities are missing. Consequently, a larger body of works focuses on the completion of missing information in KGs, which is commonly referred to as link prediction (LP). This task has traditionally and extensively been studied in the \textit{transductive} setting, where all entities and relations in the testing set are observed during training.
Recently, several works have tackled the LP task under more challenging settings, where entities and relations in the test set may be unobserved during training, or appear in only a few facts. 
These works are known as \emph{inductive}, \emph{few-shot}, and \emph{zero-shot} link prediction.
In this work, we conduct a systematic review of existing works in this area. A thorough analysis leads us to point out the undesirable existence of diverging terminologies and task definitions for the aforementioned settings, which further limits the possibility of comparison between recent works.
We consequently aim at dissecting each setting thoroughly, attempting to reveal its intrinsic characteristics. A unifying nomenclature is ultimately proposed to refer to each of them in a simple and consistent manner.
\end{abstract}

\ccsdesc[500]{Computing methodologies~Machine learning}
\ccsdesc[500]{Computing methodologies~Knowledge representation and reasoning}

\keywords{Knowledge Graph Embeddings, Link Prediction, Few-Shot Learning, Zero-Shot Learning, Inductive Reasoning}


\maketitle

\section{Introduction}
Knowledge graphs (KGs) have emerged as a prominent data representation and management paradigm.
KGs are inherently incomplete, incorrect, or overlapping, and thus major refinement tasks include entity matching and link prediction~\cite{wang2017}. The latter is the focus of this paper.
Link prediction (LP) aims at completing KGs by inferring missing facts.
In the LP task, one is provided with a set of incomplete triples, where the missing head (resp. tail) needs to be predicted. 

Mainstream approaches for performing LP rely on knowledge graph embedding models (KGEMs) due to their ability to encode the structured knowledge into low-dimensional vectors, known as \emph{embeddings}. By mapping entities and relations into a continuous vector space, these models enable efficient prediction of missing links -- usually by simple vector operations. 

Using KGEMs to perform LP has traditionally been done in the transductive setting, which assumes that all entities and relations in the graph are observed during the training phase.
Recently, there has been a growing interest in extending KGEMs beyond the transductive setting. Researchers have explored new approaches to tackle challenges such as few-shot link prediction (FSLP), zero-shot link prediction (ZSLP), and inductive link prediction (ILP).

FSLP aims at predicting missing links when only limited prior information is available for some long-tail relations.
ZSLP addresses the scenario where the target relations have no observed training examples at all.
ILP aims at making predictions for entirely new entities that were not present in the training data.

\subsection{Motivations and Contributions}
FSLP, ZSLP, and ILP are rather emerging settings compared to the transductive one~\cite{nickel2011}. In this work, we point out the lack of common understanding revolving around these tasks. 
Specifically, the contributions of our work are as follows:
\begin{itemize}
    \item To the best of our knowledge, this work is the first to systematically review the existing literature around the FSLP, ZSLP, and ILP settings. It provides a bird's-eye view of model design commonalities, datasets used for evaluation, and general trends in the field.
    \item We conduct a thorough analysis of selected papers, which clearly pinpoints a lack of rigorous conceptualization and understanding of the intrinsic characteristics of each setting. In particular, we observe that the aforementioned tasks do not have consistent definitions between papers. Consequently, we provide a terminological discussion that aims at clearing up incidental questions and providing a unifying understanding of each setting.
    \item We propose a consistent and pragmatic naming nomenclature for the full range of LP settings, which avoids any overlap between them. Transductive LP, FSLP, ZSLP, and ILP -- as the way they are currently defined -- all have a concrete mapping in the proposed nomenclature.
\end{itemize}
\subsection{Comparison to Existing Works}
Link prediction with KGEMs is a popular task which has been extensively studied and analyzed, as evidenced by the plethora of available surveys (Table~\ref{tab:surveys}). However, these surveys are mainly concerned with LP in the transductive setting. 

A few surveys are concerned with few-shot learning (FSL) and zero-shot learning (ZSL). However, they do not directly address the LP task. They consider the task as a whole, \textit{i.e.} not focusing on any particular domain~\cite{wang2019survey, wang2021, chen2021}, or on the contrary, are concerned with a particular domain of application -- \textit{e.g.} object detection~\cite{antonelli2022, huang2023} -- or a particular family of learning paradigms, such as reinforcement learning~\cite{wang2023RL}.

Few surveys actually deal with LP beyond the transductive setting (Table~\ref{tab:surveys}). Ma and Wang~\cite{ma2023} are concerned with FSLP, but restrain themselves to FSLP with commonsense KGs. The survey from Zhang \textit{et al.}~\cite{zhang2022survey} thoroughly explores the FSLP task, and focuses on models, applications and challenges for future work. However, the number of referenced papers is relatively low due to the non-systematic approach and the non-inclusion of most recent works (Table~\ref{tab:surveys}).

To the best of our knowledge, the recent survey from Chen \textit{et al.}~\cite{chen2023survey} is the only work dealing with FSLP, ZSLP, and ILP at the same time. However, they do so by unifying all these settings under a common term, \emph{knowledge extrapolation}. Differences are made at the level of model designs and datasets. The intrinsic characteristics defining FSLP, ZSLP, and ILP, are left unassessed. This severely restricts the possibility of understanding how these settings relate to each other. 

In contrast, our work aims at dissecting these settings and providing a comprehensive understanding of each of them. In addition, this work makes clear that a confusion exists around these settings due to the way they are currently defined. The lack of a commonly agreed terminology causes authors to propose distinct and sometimes conflicting definitions to refer to these settings. Therefore, the purpose of this work is also to propose a rationale for defining and clarifying them, hoping that the proposed terminology would serve as a basis for nurturing future discussions on the LP task beyond the transductive setting.

\begin{table}[t]
\centering
\caption{Comparison of surveys targeting the LP task. Topic column header indicates the main surveyed aspects: Applications (A), Model Benchmarking (B), Model design (M), Future Perspectives (P), Reproducibility (R), and Terminology (T). Systematic column header indicates whether a systematic approach on how relevant papers were retrieved and selected is presented in the paper. Settings column header is split into Inductive (I), Few-shot (FS), and Zero-shot (ZS) link prediction, where dots indicate the number of papers considered: \dashed: not covered, \lightgreenpoint: 1-10 references covered; \pastelgreenpoint: 11-40 references covered; \flashygreenpoint: 40+ references covered. The \#Ref. column header indicates the total count of papers related to these settings.}\label{tab:surveys}
\begin{tabular}{p{2.75cm}p{0.75cm}p{1.5cm}ccccccc}
\hline
Authors & Year & Topic & Terminology & Syst. & \#Ref. & \multicolumn{3}{c}{Settings} \\
\cline{7-9}
 & & & & & & I & FS & ZS \\
\hline
Wang \textit{et al.} \cite{wang2017} & 2017 & A, M & \xmark & \xmark & \dashed & \dashed & \dashed & \dashed \\
Nguyen \textit{et al.} \cite{nguyen2020survey} & 2020 & B & \xmark & \xmark & \dashed & \dashed & \dashed & \dashed \\
Dai \textit{et al.} \cite{dai2020survey} & 2020 & A, B, M, P & \xmark & \xmark & \dashed & \dashed & \dashed & \dashed \\
Ji \textit{et al.} \cite{ji2020survey} & 2020 & A, M & \xmark & \xmark & \dashed & \dashed & \dashed & \dashed \\
Chen \textit{et al.} \cite{chen2020survey} & 2020 & B, M, P & \xmark & \xmark & \dashed & \dashed & \dashed & \dashed \\
Chen \textit{et al.} \cite{chen2021survey} & 2021 & B, M, P & \xmark & \xmark & $5$ & \lightgreenpoint & \dashed & \lightgreenpoint \\
Rossi \textit{et al.} \cite{rossi2021} & 2021 & B, M & \xmark & \xmark & \dashed & \dashed & \dashed & \dashed \\
Wang \textit{et al.} \cite{wang2021survey} & 2021 & B, M & \xmark & \xmark & \dashed & \dashed & \dashed & \dashed \\
Zamini \textit{et al.} \cite{zamini2022survey} & 2022 & B, M & \xmark & \xmark & \dashed & \dashed & \dashed & \dashed \\
Ferrari \textit{et al.} \cite{ferrari2022survey} & 2022 & B, M, P, R & \xmark & \xmark & \dashed & \dashed & \dashed & \dashed \\
Zhang \textit{et al.} \cite{zhang2022survey} & 2022 & M, P & \xmark & \xmark & $16$ & \dashed & \pastelgreenpoint & \lightgreenpoint \\
Zope \textit{et al.} \cite{zope2023survey} & 2023 & R & \xmark & \xmark & \dashed & \dashed & \dashed & \dashed \\
Braken \textit{et al.} \cite{braken2023survey} & 2023 & B & \xmark & \xmark & $3$ & \dashed & \lightgreenpoint & \dashed \\
Ma \textit{et al.} \cite{ma2023} & 2023 & A, M, P & \xmark & \xmark & $27$ & \lightgreenpoint & \pastelgreenpoint & \lightgreenpoint \\
Chen \textit{et al.} \cite{chen2023surveyfslzsl} & 2023 & A, M, P & \xmark & \xmark & $29$ & \pastelgreenpoint & \lightgreenpoint & \lightgreenpoint \\
Chen \textit{et al.} \cite{chen2023survey} & 2023 & M, P & \xmark & \xmark & $48$  & \pastelgreenpoint & \lightgreenpoint & \lightgreenpoint \\
Hubert \textit{et al.} (ours) & 2023 & M, T & \cmark & \cmark & $129$  & \flashygreenpoint & \flashygreenpoint & \pastelgreenpoint \\
\hline
\end{tabular}
\end{table}

\subsection{Survey Structure}
Our survey is structured as follows.
A systematic literature review of existing work is presented in Section~\ref{sec:systematic}. A thorough analysis and discussion about task definition and terminology is conducted in Section~\ref{sec:current-state}, based on which we propose a pragmatic nomenclature mapping every LP-related setting. Future prospects and key insights are summed up in Section~\ref{sec:conclusion}.

\section{Beyond the Transductive Setting: A Systematic Survey}\label{sec:systematic}
In this section, more details are given on the methodology used for performing the systematic literature review. We elaborate on selected digital libraries, search queries, inclusion and exclusion criteria, and the general selection process for retrieved papers.
After this, each paper is presented individually w.r.t. several criteria. Is it essential to highlight that, at this stage, we do not perform any discussion on task definition and used terminology. Instead, we stick to the terminology used in the original works to classify these works the way they identify themselves. 

\subsection{Methodology}
\subsubsection{Search Engines and Queries.}
DBLP\footnote{\url{https://dblp.org/}}, Google Scholar\footnote{\url{https://scholar.google.com/}}, Scopus\footnote{\url{https://www.scopus.com/}}, and Web of Science\footnote{\url{https://www.webofknowledge.com/}} were selected as digital libraries for performing the query-based search and retrieving relevant papers.

\begin{figure}[h]
  \centering
  \includegraphics[width=0.75\textwidth]{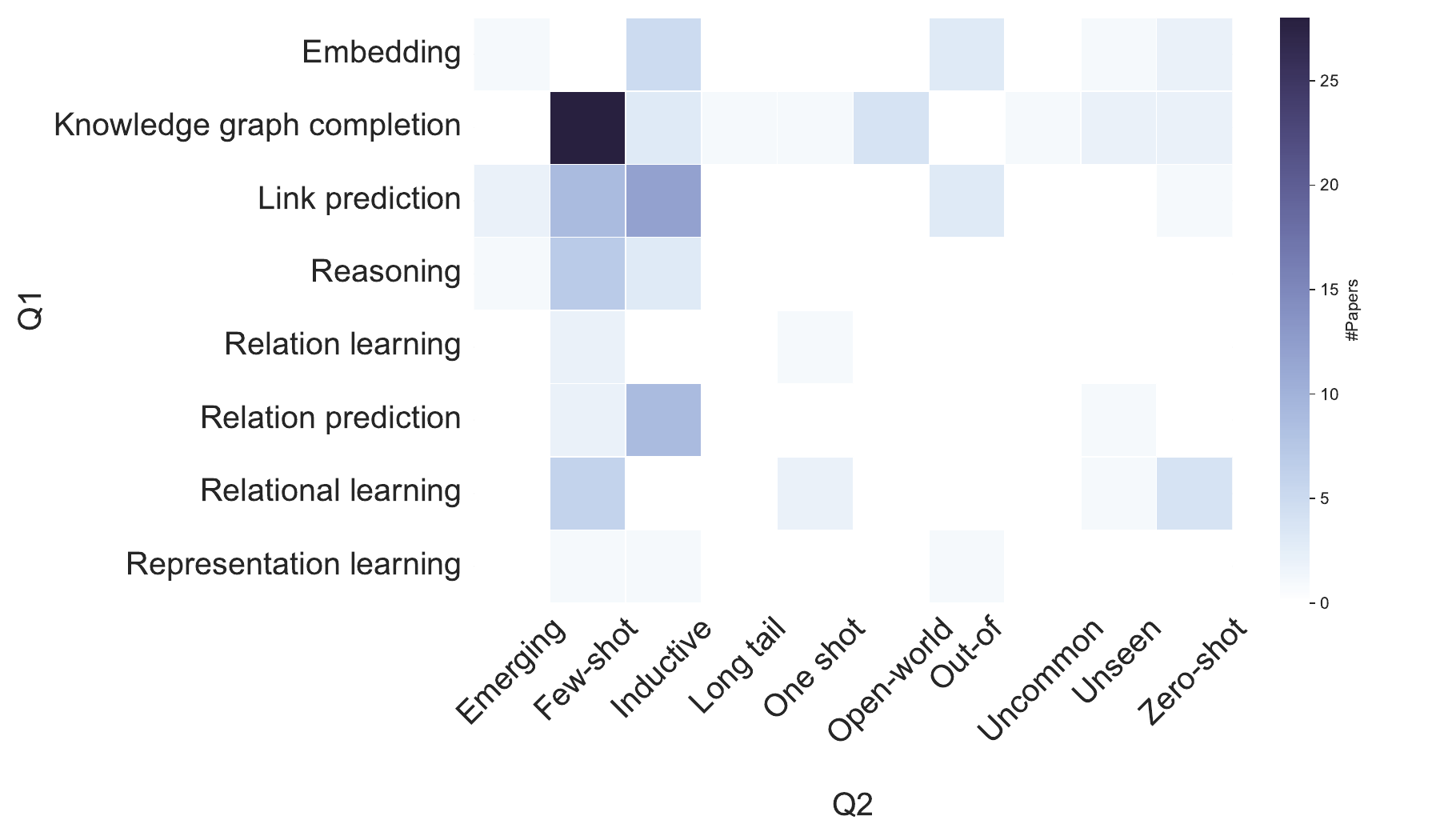}
  \caption{Association strength between each combination of keywords in Q1 and Q2 in the selected papers}
  \label{fig:heatmap}
\end{figure}

The full search query is directly scoped by the purpose of the current survey, namely analyzing papers related to LP in knowledge-scarce scenarios. As such, the query naturally consists of two sub-queries. The first one (Q1) targets the LP task, which is frequently associated with the notions of ``knowledge graph completion'', ``relation prediction'', and others (see Q1 expression in Table~\ref{tab:search-queries}). The second one (Q2) refers to the nature and degree of available information on unseen entities and relations, which are sometimes said to be ``emerging'' or ``out-of-graph'', for example. Each sub-query was incrementally refined based on retrieved papers and cross-referencing. Q1 and Q2 as they appear in Table~\ref{tab:search-queries} are the refined queries we combined to specifically target papers concerned with the LP task in knowledge-scarce scenarios. Fig.~\ref{fig:heatmap} reflects the pairwise association strength between each combination of terms from Q1 and Q2 in the 129 selected papers (see Section~\ref{sec:execution} and Fig.~\ref{fig:flowchart} for more details on the selection procedure).

\begin{table}[t]
  \centering
  \caption{Sub-queries for the full search query $Q = Q_1 \cap Q_2$}\label{tab:search-queries}
  \begin{tabular}{p{0.15\columnwidth}p{0.75\columnwidth}}
    \hline
    \textbf{Sub-Query} & \textbf{Expression} \\
    \hline
    Q1 & (``link prediction'' OR ``knowledge base completion'' OR ``knowledge graph completion'' OR ``representation learning'' OR ``relation learning'' OR ``relational learning'' OR ``reasoning'' OR ``fact prediction'' OR ``edge prediction'' OR ``triple completion'' OR ``triple prediction'' OR ``relation prediction'' OR ``embedding'' OR ``multi-relational data'') \\
    \hline
    Q2 & (``zero shot'' OR ``few shot'' OR ``K shot'' OR ``one shot'' OR ``inductive'' OR ``unseen'' OR ``uncommon'' OR ``emerging'' OR ``out of'' OR ``long tail'' OR ``open world'' OR ``scarce'' OR ``scarcity'' OR ``paucity'') \\
    \hline
  \end{tabular}
\end{table}

\subsubsection{Selection Criteria.}
In this work, eight selection criteria are considered for filtering relevant papers. For each of them we define an inclusion criterion (IC) and an exclusion criterion (EC). These are reported in Table~\ref{tab:selection-criteria}. The first five criteria (C1-C5) are automatically applied upstream, while C6-C8 require reading papers to some extent.
 
We found some references related to either FSLP, ZSLP, or ILP, that additionally revolve around such aspects such as temporal LP~\cite{bai2023fsltemporal,ding2023fslTEMPORAL}, or multi-modal LP~\cite{zhang2022fslmultimodal}. Such papers are outside the scope of the systematic part of this survey, but are briefly discussed in Section~\ref{sec:conclusion} to broaden our discussion to future prospects in the field.

\begin{table}[htbp]
  \centering
  \caption{Criteria for selecting papers}\label{tab:selection-criteria}
  \begin{tabular}{p{0.21\columnwidth}p{0.35\columnwidth}p{0.35\columnwidth}}
    \toprule
    \textbf{Criterion} & \textbf{Inclusion Criterion (IC)} & \textbf{Exclusion Criterion (EC)}\\
    \midrule
    C1 Language & English & Other than English \\
    \midrule
    C2 Period & 2017-2023 & Works published before 2017. \\
    \midrule
    C3 Publication type & Peer-reviewed articles including main conference papers, workshop papers, journal papers, and book chapters. & Non-peer-reviewed papers and/or theses, extended abstracts, supplementary materials, surveys, presentations, tutorials, and patents. \\
    \midrule
    C4 Accessibility & Papers that are accessible from a major research institution (University of Lorraine, France) without additional paywall. & Papers whose access is restricted from the University of Lorraine and that would require purchasing them. \\
    \midrule
    C5 Duplicates & In case of duplicated publications, the latest version is considered. & Papers for which a more recent version exists are not considered. \\
    \midrule
    \midrule
    C6 Manual Filtering & Papers whose title and/or abstract make clear that experiments are carried out w.r.t. link prediction in at least one non-transductive setting are considered. & Papers concerned with other tasks than link prediction, or link prediction in a transductive setting only, are not considered. \\
    \midrule
    C7 Data type & Papers should be concerned about directed, edge-labeled multi-relational graphs composed of triples. & Papers concerned with homogeneous graphs and non-graph datasets are not considered. In addition, papers concerned with hypergraphs, temporal KGs, or multi-modal KGs, are not considered in our main discussion.\\
    \midrule
    C8 Survey scope & After a full read, papers that dedicate at least part of their focus on FSLP, ZSLP, or ILP are considered. & After full read, papers that do not discuss any of FSLP, ZSLP, or ILP are not considered.\\
    \bottomrule
  \end{tabular}
\end{table}

\subsubsection{Methodology Execution.}\label{sec:execution}
\begin{figure}[h]
  \centering
  \includegraphics[width=0.95\textwidth]{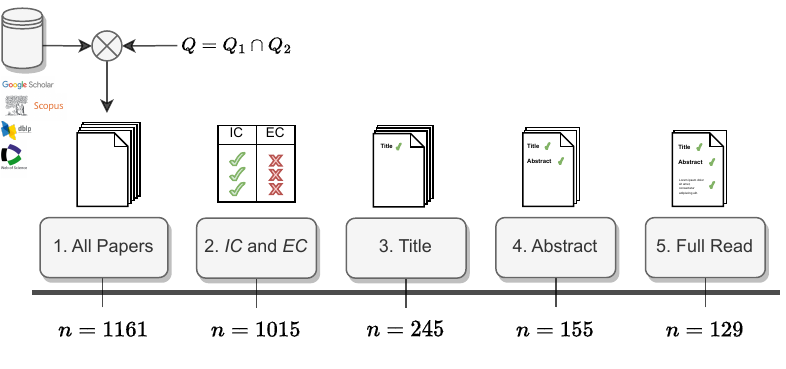}
  \caption{Systematic paper selection flowchart}
  \label{fig:flowchart}
\end{figure}

After running the search queries\footnote{The search query was last run on 30 September 2023.} in the selected digital libraries and removing duplicate papers, a total of $1,161$ unique papers was retrieved (Fig.~\ref{fig:flowchart}). Applying the aforedescribed selection criteria C1-C5, we were left with $1,015$ papers.

Manual filtering on titles and abstracts (C6) was performed afterwards, which led us to keep $245$ and then $155$ papers, respectively. After a complete reading of the $155$ remaining papers, $129$ of them which rigorously fit the scope of our survey (C7-C8) were selected for further analysis.

\subsection{General Overview}
Before discussing the FSLP, ZSLP, and ILP tasks, all works selected according to the above methodology are described indiscriminately. This aims at providing an overall picture of the field, and identifying trends regardless of any particular non-transductive setting.

It is worth mentioning that several works refer to both ILP and FSLP \cite{baek2020,yang2022fslpilp,zheng2023fslpilp,zhao2023fslpilp}. This is because some approaches draw characteristics from both FSLP and ILP works at the same time. When this combination is justified (\textit{e.g.} by presenting experiments in the two settings), such papers are considered as related to both ILP and FSLP in the current section. They are consequently reported in both Table~\ref{tab:fslp-models} and Table~\ref{tab:ilp-models}.

However, this potentially highlights inconsistencies or insufficient framing of tasks definitions in the research community. In particular, it remains to be determined whether ILP and FSLP are two disjoint tasks. This discussion is presented in Section~\ref{sec:current-state}, where the focus is on clearly understanding the peculiarities of ILP, FSLP, and ZSLP, and subsequently presenting a unifying and comprehensive framework to refer to them unambiguously.

\begin{figure}[t]
  \centering
  \includegraphics[width=0.55\textwidth]{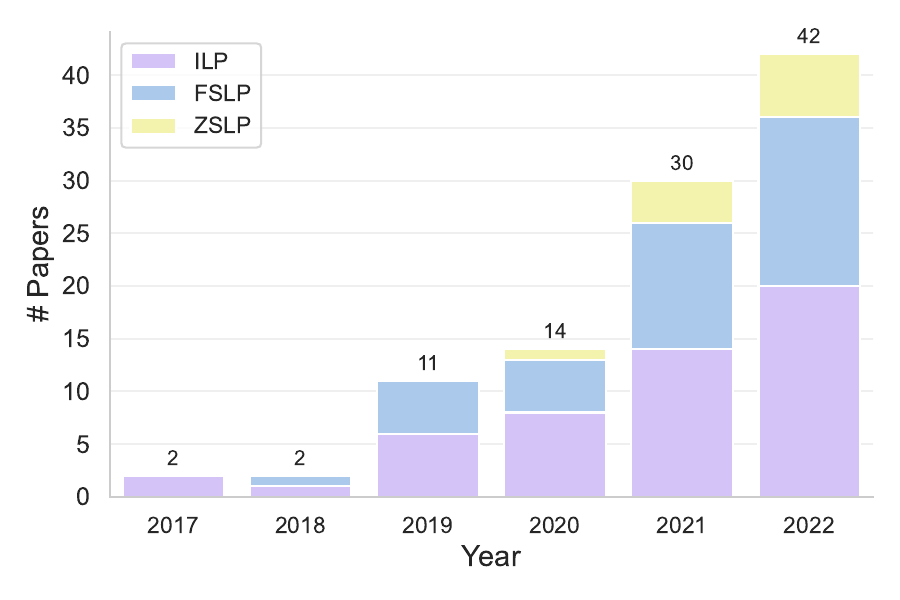}
  \caption{Paper count over the years}
  \label{fig:paper-count}
\end{figure}

\begin{figure}[t]
  \centering
  \includegraphics[width=0.55\textwidth, trim=0.25cm 1.75cm 0.25cm 0.5cm, clip]{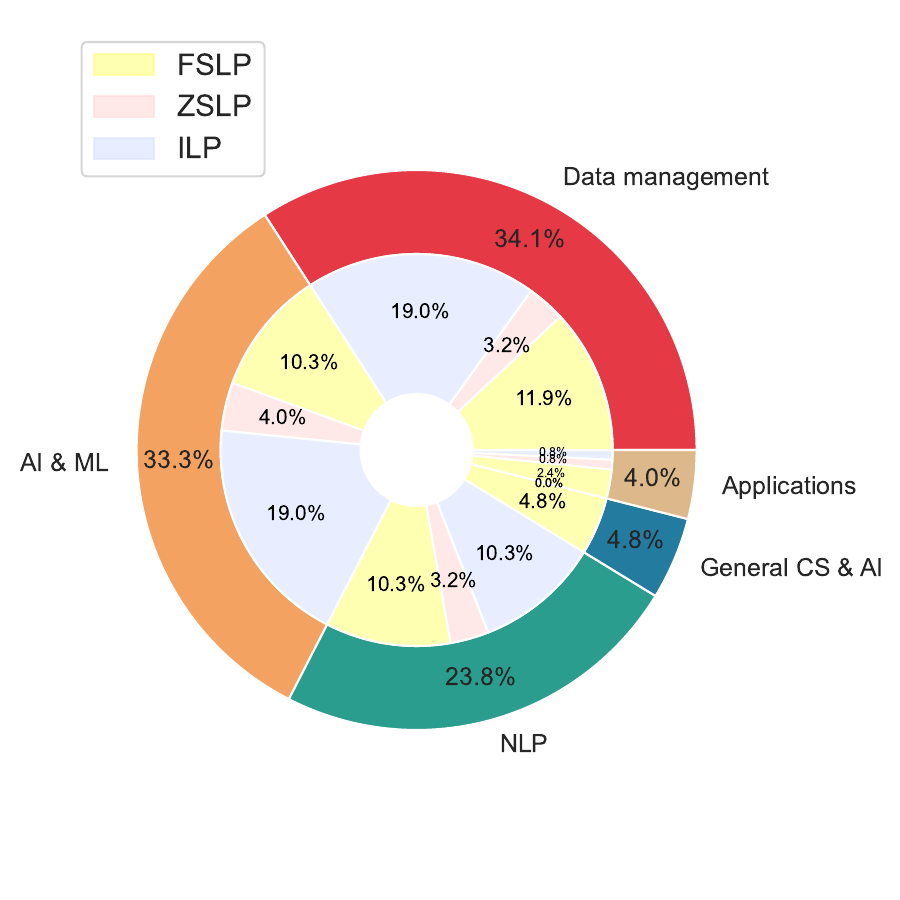}
  \caption{Main domains for selected papers}
  \label{fig:pie-chart-venues}
\end{figure}

Based on the collection of selected papers, it is obvious that LP beyond the transductive setting is gaining momentum in terms of research interest: the number of published works related to these settings steadily increases over the years (Fig.~\ref{fig:paper-count}). This research interest spans over multiple venues with initially different topics and expertise: Fig~\ref{fig:pie-chart-venues} groups publications by general domains. Specifically, selected papers can be published in data management venues\footnote{ACM Trans. Inf. Syst. (Journal), ACM Trans. Web (Journal), BDEIM, Big Data Research (Journal), CIKM, DASFAA, Expert Systems With Applications (Journal), ICDE, IJCKG, Inf. Sci. (Journal), ISWC, KDD, KSEM, PAKDD, SDM, SIGIR, TKDE (Journal), WSDM, WWW.}, AI and ML-focused venues\footnote{AAAI, ECML/PKDD, ICCPR, ICLR, ICML, ICTAI, IJCAI, IJCNN, Artif. Int. Res. (Journal), Neural Networks (Journal), NeurIPS, Neurocomputing (Journal), Pattern Recognit. Lett. (Journal).}, NLP venues\footnote{ACL, COLING, EMNLP, LREC, NAACL, Trans. Asian Low-Resour. Lang. Inf. Process. (Journal).}, general CS and AI venues\footnote{Applied Sciences (Journal), Electronics (Journal), IEEE Access (Journal), SAC, SIGAPP Appl. Comput. Rev. (Journal).} and venues focused on applications, e.g., in bioinformatics, energy, networks and other fields\footnote{BIBM, EI2, ICASSP, ICIIP.}. However, not all settings are equally investigated: ILP and FSLP are by far the most studied settings, while roughly 11\% of selected papers (13 out of 129) identify themselves as concerned with ZSLP, the vast majority of which coming from the years 2021-2022. Moreover, we observe a non-uniform distribution of approaches across venues.

In what follows, the three studied settings are defined according to their predominant definitions as found in relevant papers. Then, works that relate to these settings are thoroughly presented and compared. 

\subsection{Few-Shot Learning Link Prediction}\label{sec:fslp}

\subsubsection{Task Definition} 

Few-shot Learning (FSL) was first introduced in the context of learning a classifier from a single
example~\cite{fink2004,feifei2006}, \textit{i.e.} one-shot learning (OSL). As an extension to larger K-shot settings, FSL encompasses OSL, provided that $K$ remains low. Wang \textit{et al.}~\cite{wang2021} define FSL as a learning problem described by an experience (E) and a set of tasks (T), where E contains only a limited number of examples with supervised information for the target T. Under the FSL paradigm, classes covered by training instances and the ground-truth classes for testing instances are disjoint. However, learning classifiers for the classes of testing instances is usually performed by utilizing knowledge contained in instances of other related classes, observed during training.

The FSL problem spans across image classification~\cite{vinyals2016}, object recognition~\cite{feifei2006}, and sentiment classification~\cite{yu2018}.
It has recently been extending to the LP task. FSLP and K-shot LP are formally defined as follows:
\vspace{0.15cm}
\begin{definition}
    [Few-Shot Link Prediction] In addition to a background knowledge graph $\mathcal{KG} = \{ \mathcal{E}, \mathcal{R}, \mathcal{T}\}$ on which a model has been pre-trained, we are given a new relation $r' \notin \mathcal{R}$, along with its support set  $\mathcal{S}_{r'} = \{(h, t)\in \mathcal{E} \times \mathcal{E} | (h, r', t) \notin \mathcal{T} \}$ with $|\mathcal{S}_{r'}|$ being small. FSLP consists in predicting the tail entity linked with relation $r'$ to head entity $h$, considering a candidate set of tail entities $C_{(h,r')}$.
\end{definition}
\vspace{0.15cm}
\begin{definition}
    [K-Shot Link Prediction] On the basis of the above FSLP definition, when $|\mathcal{S}_{r'}| = K$ and $K << |\mathcal{S}_r|$ for all $r \in \mathcal{KG}$, the task is called K-shot LP.
\end{definition}
\vspace{0.15cm}
Thus, the goal of FSLP is to learn a model able to predict the right missing entities in triples of a relation $r'$, with only observing a few triples of $r'$. These few triples form the \emph{support set} $\mathcal{S}_r'$ and act as supportive knowledge for the model to learn, whereas the \emph{query set} $\mathcal{Q}_r' = \{ r':(h_j, ?)\}$ contains triples to predict.

In the few-shot learning setting, training is performed w.r.t. a set of tasks $\mathbb{T}_{train}=\{\mathbb{T}_{r_{i}}\}_{i=1}^{M}$. Each task -- a.k.a. episode~\cite{vinyals2016} -- is concerned with one single relation $r \in \mathcal{R}$, which means that all triples of the relation $r$ are bundled together in the same training task $\mathbb{T}_{r}$, where each task $\mathbb{T}_{r_{i}} = \{\mathcal{S}_{r_{i}}, \mathcal{Q}_{r_{i}}\}$ has its own support and query set.
Ultimately, testing is performed on a set of new tasks $\mathbb{T}_{test} = \{\mathbb{T}_{r_{j}}\}_{j=M+1}^{N}$ following the exact same procedure as with training tasks, except that the set of relations ${r_{j}} \in \mathbb{T}_{test}$ is disjoint with relations ${r_{i}} \in \mathbb{T}_{train}$. Notably, predictions are performed by considering a candidate set of tail entities $C_{(h,r)}$ w.r.t. to a query $(h,r)$. This means that only a subset of entities are scored and ranked -- more specifically those entities that are semantically valid with respect to the relation $r$ at hand~\cite{xiong2018}.
\vspace{0.15cm}

\begin{table}[htbp]
\centering
\scriptsize
\caption{Presentation of FSLP works. The last four column headers refer to the incorporation of external knowledge, the learning paradigm, the nature of the unseen element (entity and/or relation), and the nature of the element to predict. Dashed lines materialize the absence or non-relevancy of certain information. Regarding the model column, some authors do not necessarily propose their own model. Regarding external knowledge, dashed lines also denote the absence of such information being used. Regarding the paradigm, Yao \textit{et al.} \cite{yao2022fsl} propose a data augmentation technique which is agnostic to any learning paradigm, hence the absence of information for this field}\label{tab:fslp-models}
\begin{tabular}{p{2.25cm}p{0.5cm}p{2.75cm}p{2cm}p{3.25cm}p{0.5cm}p{0.25cm}}
\hline
Authors & Year & Model & Ext. Knowl. & Paradigm & U & P \\
\hline
Huang \textit{et al.} \cite{huang2022fsl} & 2018 & CSR & \dashed & Metric-based & R & E \\
Xiong \textit{et al.} \cite{xiong2018} & 2018 & GMatching & \dashed & Metric-based & R & E \\
Xie \textit{et al.} \cite{xie2019fsl} & 2019 & \dashed & \dashed & Metric-based & R & E \\
Baek \textit{et al.}\footnote{This work is concerned with both ILP and FSLP. Consequently, it is reported in the corresponding tables.} \cite{baek2020} & 2020 & GEN & \dashed & Metric-based & E & E \\
Ioannidis \textit{et al.} \cite{ioannidis2020fsl} & 2020 & I-RGCN & \dashed & Metric-based & R & E \\
Sheng \textit{et al.} \cite{sheng2020fsl} & 2020 & FAAN & \dashed & Metric-based & R & E \\
Zhang \textit{et al.} \cite{zhang2020fsl} & 2020 & FSRL & \dashed & Metric-based & R & E \\
Sun \textit{et al.} \cite{sun2021fsl} & 2021 & Att-Metric & \dashed & Metric-based & R & R \\
Wang \textit{et al.} \cite{wang2021fslGNN} & 2021 & \dashed & \dashed & Metric-based & R & E \\
Wang \textit{et al.} \cite{wang2021fslBERT} & 2021 & B-GMatching & Text & Metric-based & R & E \\
Wang \textit{et al.} \cite{wang2021fslREFORM} & 2021 & REFORM & \dashed & Metric-based & R & E \\
Xiao \textit{et al.} \cite{xiao2021fsl} & 2021 & HMNet & \dashed & Metric-based & R & E \\
Xu \textit{et al.} \cite{xu2021fsl} & 2021 & P-INT & \dashed & Metric-based & R & E \\
Zhang \textit{et al.} \cite{zhang2021fslzsl} & 2021 & GRL & \dashed & Metric-based & R & E \\
Cornell \textit{et al.} \cite{cornell2022} & 2022 & \dashed & Text & Metric-based & R & E \\
Li \textit{et al.} \cite{li2022fslEMNLP} & 2022 & CIAN & \dashed & Metric-based & R & E \\
Li \textit{et al.} \cite{li2022fsl} & 2022 & FAAN-G & Attributes, Text & Metric-based & R & E \\
Liang \textit{et al.} \cite{liang2022fsl} & 2022 & YANA & \dashed & Metric-based & R & E \\
Ma \textit{et al.} \cite{ma2022fsl} & 2022 & AAF & \dashed & Metric-based & R & E \\
Wu \textit{et al.} \cite{wu2022fsl} & 2022 & MFEN & \dashed & Metric-based & R & E \\
Wu \textit{et al.} \cite{wu2022fsldiscriminative} & 2022 & FSDR & \dashed & Metric-based & R & E \\
Yuan \textit{et al.} \cite{yuan2022fslSAC} & 2022 & HARV & \dashed & Metric-based & R & E \\
Yuan \textit{et al.} \cite{yuan2022fsl} & 2022 & IDEAL & \dashed & Metric-based & R & E \\
Li \textit{et al.} \cite{li2023fslmpi} & 2023 & FRL-KGC & \dashed & Metric-based & R & E \\
Li \textit{et al.} \cite{li2023fslcapsule} & 2023 & InforMix-FKGC & \dashed & Metric-based & R & E \\
Liang \textit{et al.} \cite{liang2023fsltransam} & 2023 & TransAM & \dashed & Metric-based & R & E \\
Luo \textit{et al.} \cite{luo2023fslsemantic} & 2023 & SIM & \dashed & Metric-based & R & E \\
Chen \textit{et al.} \cite{chen2019metaR} & 2019 & MetaR & \dashed & Optimization-based & R & E \\
Lv \textit{et al.} \cite{lv2019fsl} & 2019 & Meta-KGR & \dashed & Optimization-based & R & E \\
Wang \textit{et al.} \cite{wang2019fsllongtail} & 2019 & TCVAE & Text & Optimization-based & E, R & E \\
Jonathan \textit{et al.} \cite{jonathan2022fsl} & 2020 & Intra-graph & \dashed & Optimization-based & R & E \\
Zhang \textit{et al.} \cite{zhang2020fslEMNLP} & 2020 & FIRE & \dashed & Optimization-based & R & E \\
Jambor \textit{et al.} \cite{jambor2021fsl} & 2021 & ShareEmbed, ZeroShot & \dashed & Optimization-based & R & E \\
Jiang \textit{et al.} \cite{jiang2021fsl} & 2021 & MetaP & \dashed & Optimization-based & R & E \\
Niu \textit{et al.} \cite{niu2021fsl} & 2021 & GANA-MTransH & \dashed & Optimization-based & R & E \\
Zheng \textit{et al.} \cite{zheng2021fslmetaRL} & 2021 & THML & \dashed & Optimization-based & E & E \\
He \textit{et al.} \cite{he2022fsl} & 2022 & ERRM & \dashed & Optimization-based & R & E \\
Yang \textit{et al.} \cite{yang2022fslpilp} & 2022 & MILP & \dashed & Optimization-based & E & E \\
Ye \textit{et al.} \cite{ye2022fsl} & 2022 & OntoPrompt & Ontology, Text & Optimization-based & R & E \\
Zhang \textit{et al.} \cite{zhang2022fsl} & 2022 & ADK-KG & Text & Optimization-based & R & E \\
Cai \textit{et al.} \cite{cai2023fsl} & 2023 & DARL & \dashed & Optimization-based & R & E \\
Li \textit{et al.} \cite{li2023fsltransd} & 2023 & TDML & \dashed & Optimization-based & R & E \\
Pei \textit{et al.} \cite{pei2023fsl} & 2023 & FLow-KGC & \dashed & Optimization-based & R & E \\
Pei \textit{et al.} \cite{pei2023fslKDD} & 2023 & FLow-MV & \dashed & Optimization-based & R & E \\
Qiao \textit{et al.} \cite{qiao2023fsl} & 2023 & RANA & \dashed & Optimization-based & R & E \\
Wu \textit{et al.} \cite{wu2023fsl} & 2023 & HiRE & \dashed & Optimization-based & R & E \\
Zhao \textit{et al.} \cite{zhao2023fslBayesKGR} & 2023 & BayesKGR & \dashed & Optimization-based & E & E \\
Zheng \textit{et al.} \cite{zheng2023fslpilp} & 2023 & Meta-iKG & \dashed & Optimization-based & R & E \\
Du \textit{et al.} \cite{du2019fsl} & 2019 & CogKR & \dashed & DP theory& R & E \\
Luo \textit{et al.} \cite{luo2023fsl} & 2023 & NP-FKGC & \dashed & Neural process & R & E \\
Zhao \textit{et al.} \cite{zhao2023fslpilp} & 2023 & RawNP & \dashed & Neural process & E & E \\
Yao \textit{et al.} \cite{yao2022fsl} & 2022 & \dashed & \dashed & \dashed & R & E \\
\hline
\end{tabular}
\end{table}

\subsubsection{Model Presentations}
Most FSLP approaches fall into two main categories, \textit{i.e.},  metric-based and optimization-based approaches~\cite{xiong2018}. However, approaches relying on the dual-process~\cite{du2019fsl} and neural process theories~\cite{luo2023fsl, zhao2023fslpilp} are also observed in recent works (Table~\ref{tab:fslp-models}).

\sstitle{Metric-based} models usually encode head and tail entity pairs of an unseen relation and then match these encoded entity pairs with entity pairs in the query set to predict whether they are connected by the same unseen relation.

GMatching~\cite{xiong2018} is the first metric-based model proposed for the FSLP task, and its performance has originally been evaluated w.r.t. one-shot LP. GMatching produces entity embeddings by looking at the local graph structure of entities and learns a matching metric to assess the similarity between queries and the support triples. 
Incremental improvements of GMatching are proposed by Xie \textit{et al.}~\cite{xie2019fsl}, who add an attention mechanism to GMatching in order to improve the quality of the learnt embeddings.
Also targeting the one-shot LP task, Sun \textit{et al.}~\cite{sun2021fsl} propose a framework based on attention neighborhood aggregation and paths encoding to better predict on low-frequency relations.

While GMatching was concerned with one-shot LP, FSRL~\cite{zhang2020fsl} builds on the neighbor information aggregation mechanism of GMatching and extends it to the few-shot setting. FSRL further considers information from one-hop entity neighbors using a fixed attention mechanism.
Sheng \textit{et al.} propose an extension named FAAN~\cite{sheng2020fsl}: although their model also relies on the attention mechanism, it does so by applying an adaptive attentional neighbor encoder, which allows entity representations to be based on the semantic role they play w.r.t. different relations. Equipped with an additional method of data enhancement which increases data quantity, FAAN-G~\cite{li2022fsl} appears as an improvement over the initial FAAN model.

This idea of variable attention weights based on neighboring entities led to many more models tackling the FSLP task, e.g., AAF~\cite{ma2022fsl} and IDEAL~\cite{yuan2022fsl}, which exploit hiearchical attention mechanisms, MFEN~\cite{wu2022fsl}, which is based on heterogeneous representation learning, capturing the heterogeneous roles of the relational neighbors of a given entity and all of their features via a convolutional encoder. FSD~\cite{wu2022fsldiscriminative} integrates negative support triples in its overall training framework, which leads to better interaction modeling between entities. In addition, while many existing methods treat the support set as a collection of independent triples, FSDR models the interactions among support triples.

HARV~\cite{yuan2022fslSAC} aims at capturing the differences between neighbor relations and entities and interaction between relations, which were left unconsidered in previous works. 
Similarly, CIAN~\cite{li2022fslEMNLP} exploits the interactions within entities by computing the attention between the task relation and each entity neighbor.

Ioannidis \textit{et al.}~\cite{ioannidis2020fsl} note that GNNs are not good at learning relation embeddings for unseen or low-frequency relations. Therefore, they propose inductive RGCN (I-RGCN) for learning informative relation embeddings even in the few-shot setting.
GEN~\cite{baek2020} is a model that learns to extrapolate the knowledge of a given graph to unseen entities. Equipped with a stochastic
transductive layer, GEN is not only able to propagate knowledge between seen and unseen entities, but also between unseen entities.
Also facing current limitations of GNNs for modeling low-frequency relations, Wang \textit{et al.}~\cite{wang2021fslGNN} propose a GNN-based model that leverages intra-layer neighborhood attention and inter-layer memory attention, which alleviates over-smoothing tendencies of vanilla GNNs.
YANA~\cite{liang2022fsl} constructs a local pattern graph from entities, based on which a modified version of R\-GCN is applied to capture hidden dependencies between entities. In addition, a query-aware gating mechanism combines topology signals from the local pattern graph with semantic information learned from a given entity neighborhood.

HMNet~\cite{xiao2021fsl} is a hybrid matching network that evaluates triples from both entity and relation perspectives. At the entity-aware matching network, HMNet uses an attentive embedding layer to aggregate entity features and relation-aware topology. At the relation-aware matching network, it leverages a feature attention mechanism to implement the relation perspective evaluation.

REFORM~\cite{wang2021fslREFORM} focuses on error-aware few-shot LP and is based on three modules: a neighbor encoder, a cross-relation aggregation, and an error mitigation module. REFORM is reported to beat state-of-the-art models such as GMatching, FAAN, and FSRL, on few-shot benchmarks.

Huang \textit{et al.} propose CSR~\cite{huang2022fsl}, a model that uses connection subgraphs among the support triples and tests whether they are connected to the query triples.

Based on the expressiveness of path-based information, P-INT~\cite{xu2021fsl} tackles the FSLP problem by building the interactions of paths between support and query entity pairs, which helps capture finer-grained matches between entity pairs.

Regarding metric-based models, the transformer framework is increasingly used as a way to enhance entity representation with capturing semantic interaction between entity neighbors~\cite{liang2023fsltransam,li2023fslmpi,luo2023fslsemantic}.
FRL-KGC~\cite{li2023fslmpi} uses a gating mechanism to better aggregate distant neighbors and avoid noisy signals to propagate from them, and exploits correlations between entity pairs in the reference set for a more accurate relation representations. TransAM~\cite{liang2023fsltransam} also explores the interactions between entities by leveraging the co-occurrence of two entities. 
SIM~\cite{luo2023fslsemantic} uses an entity-relation fusion module to adaptively encode neighbors while incorporating relation representation.

\sstitle{Optimization-based} models are generally parameterized by meta parameters, which are aimed at being updated quickly via gradients on few-shot samples, so that the model can seamlessly generalize to new tasks. Optimization-based models may rely on Model Agnostic Meta-Learning (MAML) for fast adaption of relation-specific meta information from the embeddings of entities in the support set and transferring it to the query set.

MetaR~\cite{chen2019metaR} is the first representative work based on this idea. It works by extracting relation-specific meta information and uses it for few-shot relational predictions.
By the same token, MetaP~\cite{jiang2021fsl} uses a meta-based pattern extracting framework, in which the plausibility of triples  is directly linked to the relation-specific patterns.
GANA~\cite{niu2021fsl} builds on the firstly introduced model -- MetaR -- and extends it by adding an attention mechanism and a Long Short-Term Memory (LSTM) aggregator to improve the quality of generated embeddings and the relation meta representation.
By proposing to jointly capture three levels of relational information, namely entity-level, triple-level, and context-level information, HiRE~\cite{wu2023fsl} is often considered as an extension of GANA, using contrastive loss as an optimization target.

The recently proposed Meta-iKG~\cite{zheng2023fslpilp} model uses local subgraphs to transfer subgraph-specific information and quickly learn transferable patterns with meta-gradients.
Jonathan \textit{et al.} propose their Intra-Graph framework~\cite{jonathan2022fsl}, which leverages the most salient information such as symmetry between candidate entities of support set and query set.

Similar to the objective of the metric-based model FSDR (see above), RANA~\cite{yao2022fsl}
utilizes negative samples by assigning variable weights to negative samples and using MAML for parameter optimization for few-shot relations.

Flow-KGC~\cite{pei2023fsl} is a recently proposed model combining a few-shot learner, a task generator, and a task selector, which allows to better optimize the few-shot learner using the selected few-shot tasks. 
Similarly, Pei \textit{et al.} propose Flow-MV~\cite{pei2023fslKDD}, combining a few-shot learner, a perturbed few-shot learner, a relation knowledge distiller, and a pairwise contrastive distiller.

Meta-KGR~\cite{lv2019fsl} uses the reinforcement learning (RL) paradigm to model the multi-hop reasoning process, where a recurrent neural network (RNN) encodes the search path. Then, MAML is adopted to update model parameters from high-frequency relations and quickly adapt to few-shot relations.
FIRE~\cite{zhang2020fslEMNLP} extends Meta-KGR with a heterogeneous neighbor aggregator and a search space pruning strategy.
ADK-KG~\cite{zhang2022fsl} augments FIRE with a text-enhanced heterogeneous GNN that encode node embeddings.
Heterogeneity also stems from the roles an entity plays w.r.t. different relations. Instead of using static embeddings to encode entities, 
TDML~\cite{li2023fsltransd} is based on the intuition that entities convey different semantics within distinct few-shot episodes, and introduces relation-specific entity embeddings.
Another model based on Meta-KGR and FIRE is THML~\cite{zheng2021fslmetaRL}, which aims at finding an optimal ordering of training batches based on the hardness of relations.

In order to get fine-grained meta representations from meta knowledge, DARL~\cite{cai2023fsl} uses a dynamic neighbor encoder to make entity embeddings depend on neighboring relations.
BayesKGR~\cite{zhao2023fslBayesKGR} proposes a Bayesian inductive reasoning method and also incorporates meta-learning techniques for FSLP, which allows to extrapolate meta-knowledge derived from the background KG to emerging entities.

Other optimization-based models also use external sources of knowledge in addition to structural information. This is the case of TCVAE~\cite{wang2019fsllongtail} which uses relation textual descriptions and generates extra triples during the training stage to improve LP performance on few-shot tasks.

In addition to text-based information, OntoPrompt~\cite{ye2022fsl} also leverages ontologies and jointly optimizes text-based and ontology-based representations.

\sstitle{Other} models deviate from the almost binary opposition between metric-based and optimization-based models. 

CogKR~\cite{du2019fsl} relies on a summary and a reasoning module that work jointly to tackle the one-shot LP problem.
Noting that current metric-based and optimization-based approaches often suffer from the out-of-distribution, overfitting, and uncertainty in predictions issues, Luo \textit{et al.} propose a normalizing flow-based neural process to tackle them. Notably, their model NP-FKGC integrates a stochastic ManifoldE decoder to incorporate the neural process and handle complex few-shot relations. Similarly, RawNP~\cite{zhao2023fslpilp} is a  neural process-based method able to estimate the uncertainty when making predictions. It is based on a relational anonymous walk to extract a series of relational patterns from few-shot observations.

Yao \textit{et al.}~\cite{yao2022fsl} use a data augmentation technique from two perspectives -- inter-task and intra-task views -- which can be used to augment different models.

From a higher perspective and still irrespective to families of models, Jambor \textit{et al.}~\cite{jambor2021fsl} aim at probing the limits of the FSLP setting. More precisely, their work covers both FSLP and ZSLP as they find that a simple zero-shot baseline that does not use any relation-specific information achieves promising results.

\begin{table}[t]
\caption{Statistics of datasets used in FSLP. Column headers from left to right: number of entities, relations, triples, number of tasks (in train/dev/test splits), K-shot, original source of the dataset, and references to those papers. In particular, the K-shot column indicates the full set of K values chosen by retrieved papers. Dashed elements denote missing information (\textit{e.g.} not specified by the original authors). When different datasets are mentioned using the same name, symbols are attached as superscripts (\textit{e.g.} for FB15k-237)}\label{tab:fslp-datasets}
	\centering
    \scriptsize
	\begin{tabular}{p{0.1\columnwidth}p{0.05\columnwidth}p{0.03\columnwidth}p{0.05\columnwidth}p{0.07\columnwidth}p{0.1\columnwidth}p{0.05\columnwidth}p{0.3\columnwidth}}
		\toprule
		Dataset  & \#Ent. & \#Rel. & \#Triples & \#Tasks & $K$ & Source & References\\
		\midrule
        FR-DB & 13,176 & 178 & 49,015 & 25/5/10 & 1, 3 & \cite{pei2023fsl} & \cite{pei2023fsl, pei2023fslKDD} \\
        ES-DB & 12,382 & 144 & 54,066 & 30/5/10 & 1, 3 & \cite{pei2023fsl} & \cite{pei2023fsl, pei2023fslKDD} \\
        JA-DB & 11,805 & 128 & 28,774 & 24/4/8 & 1, 3 & \cite{pei2023fsl} & \cite{pei2023fsl, pei2023fslKDD} \\
        EL-DB & 5,231 & 111 & 13,839 & 21/4/8 & 1, 3 & \cite{pei2023fsl} & \cite{pei2023fsl, pei2023fslKDD} \\
        DBPtext & 51,768 & 319 & \dashed & 220/30/69 & 1, 4 & \cite{wang2019fsllongtail} & \cite{wang2019fsllongtail} \\
        WDtext & 60,304 & 178 & \dashed & 130/16/32 & 1, 4 & \cite{wang2019fsllongtail} & \cite{wang2019fsllongtail} \\
        NL27K & 27,221 & 404 & 175,412 & 101/13/20 & 3 & \cite{zhang2021fsluncertainty} & \cite{zhang2021fsluncertainty} \\
        COVID19 & 4,800 & 18 & 6,872 & 9/5/4 & 1 & \cite{jiang2021fsl} & \cite{jiang2021fsl} \\
        DBLP & 26,128 & 12 & 119,783 & \dashed & 10, 50, 100, 1K & \cite{ioannidis2020fsl} &  \cite{ioannidis2020fsl}\\
        IMDB & 11,616 & 12 & 17,106 & \dashed & 10, 50, 100, 1K & \cite{ioannidis2020fsl} &  \cite{ioannidis2020fsl}\\
        FB15k-One & 14,950 & 1212 & 508,791 & 53/5/10 & 1, 5 & \cite{sun2021fsl} & \cite{sun2021fsl, zhang2022fsl} \\
        $\text{FB15k-237}^{\dagger}$ & 14,478 & 237 & 309,621 & 32/5/8 & 1, 5 & \cite{xu2021fsl} &  \cite{xu2021fsl, wu2022fsldiscriminative} \\
        $\text{FB15k-237}^{\heartsuit}$ & 14,448 & 237 & 268,039 & \dashed & 1, 5, 10 & \cite{lv2019fsl} &  \cite{lv2019fsl, zhang2020fslEMNLP, huang2022fsl} \\
       $\text{FB15k-237}^{\clubsuit}$ & 14,541 & 237 & 281,624 & 75/11/33 & 5 & \cite{wang2021fslREFORM} & \cite{wang2021fslREFORM, luo2023fsl, zhao2023fslpilp}\\
        NELL-995 & 75,492 & 200 & 154,213 & \dashed & 1, 3 & \cite{lv2019fsl} & \cite{lv2019fsl, zhang2020fslEMNLP, zhao2023fslpilp} \\
        $\text{NELL}^{\dagger}$ & 68,545 & 358 & 181,109 & 40/5/22 & 5 & \cite{wang2021fslREFORM} & \cite{wang2021fslREFORM}\\
        $\text{NELL}^{\heartsuit}$ & 68,544 & 291 & 181,109 & \dashed & 3 & \cite{huang2022fsl} & \cite{huang2022fsl}\\
		$\text{NELL-One}^{\dagger}$ & 68,545 & 358 & 181,109 & 51/5/11 & 1, 5 & \cite{xiong2018} & \cite{xiong2018, du2019fsl, jiang2021fsl, sheng2020fsl, sun2021fsl, wang2021fslREFORM, cornell2022, luo2023fsl, jambor2021fsl, niu2021fsl, chen2019metaR, sheng2020fsl, zhang2020fsl, yuan2022fsl, xu2021fsl, wang2021fslBERT, xie2019fsl, xiao2021fsl, wang2021fslGNN, wu2022fsldiscriminative, he2022fsl, zhang2022fsl, ma2022fsl, li2022fsl, wu2022fsl, yuan2022fslSAC, liang2022fsl, li2023fslcapsule, qiao2023fsl, luo2023fslsemantic, liang2023fsltransam, li2023fsltransd, cai2023fsl, li2023fslmpi, li2022fslEMNLP, zhang2021fslzsl, jonathan2022fsl, yao2022fsl, wu2023fsl}\\
        Wiki & 4,838,244 & 822 & 5,859,240 & 156/16/11 & 5 & \cite{wang2021fslREFORM} & \cite{wang2021fslREFORM}\\
		Wiki-One & 4,838,244 & 822 & 5,859,240 &  133/16/34 & 1, 5 & \cite{xiong2018} & \cite{xiong2018, du2019fsl, sun2021fsl, sheng2020fsl, wang2021fslREFORM, cornell2022, luo2023fsl, jambor2021fsl, niu2021fsl, chen2019metaR, sheng2020fsl, zhang2020fsl, yuan2022fsl, wang2021fslBERT, xie2019fsl, xiao2021fsl, wang2021fslGNN, he2022fsl, zhang2022fsl, li2022fsl, wu2022fsl, yuan2022fslSAC, liang2022fsl, li2023fslcapsule, qiao2023fsl, luo2023fslsemantic, liang2023fsltransam, li2023fsltransd, cai2023fsl, li2023fslmpi, li2022fslEMNLP, jonathan2022fsl, yao2022fsl, wu2023fsl}\\
		\bottomrule
	\end{tabular}
\end{table}

\subsection{Zero-Shot Learning Link Prediction}\label{sec:zslp}
\subsubsection{Task Definition.} Early works introducing the zero-shot learning paradigm~\cite{larochelle2008,palatucci2009,lampert2009} were concerned about learning a classifier on ${D}_{train}$ -- containing samples labeled w.r.t. a set of classes $y \in {Y}_s$, and running inference on a new set of classes ${Y}_u$. 
Without any training data about classes in ${Y}_u$, knowledge derived from ${Y}_s$ is utilized to introduce a coupling between ${Y}_s$ and ${Y}_u$. Compared to FSL, ZSL is a more challenging task.

Traditionally, Standard ZSL is distinguished from Generalized ZSL. In the former case, it is assumed that only zero-shot samples are given during the testing phase, \textit{i.e.} samples from unseen classes only (${Y}_s \cap {Y}_u = \emptyset$) Generalized ZSL relaxes this constraint and allows samples from both seen and unseen classes to appear at testing time ($Y_s \subset Y_u$). 

Similarly to FSL, the ZSL scenario has been first investigated in the image classification context before being transposed to LP:
\vspace{0.15cm}
\begin{definition}[Zero-shot Link Prediction]\label{def:zslp} In ZSLP, we are given two disjoint sets of relations: $\mathcal{R}_s$ and $\mathcal{R}_u$, where $\mathcal{R}_s \cap \mathcal{R}_u =\emptyset$. $\mathcal{R}_s$ is the set of relations observed during training, whereas  $\mathcal{R}_u$ is the set of relations in the test set. Therefore, the training set is defined as ${D}_{train}=\{(h,r_{s},t)|h,t \in \mathcal{E}, r_{s} \in \mathcal{R}_{s}\}$ and the test set is defined as ${D}_{test}=\{(h,r_{u},t,\mathcal{C}_{(h,r_{u})})|h, t \in \mathcal{E}, \mathcal{C}_{(h,r_{u})} \subseteq \mathcal{E}, r_{u} \in \mathcal{R}_{u}\}$, where $t$ is the ground-truth tail entity and $\mathcal{C}_{(h,r_{u})}$ denotes a candidate set corresponding to a query $(h,r_{u})$. Given a query $(h, r_u)$, the ZSLP problem consists in finding the ground-truth tail entities $t$ from the candidate set $\mathcal{C}_{(h,r_{u})}$.
\end{definition}
\vspace{0.15cm}
Similarly to FSLP, prediction is performed on tail entities, but batches of triples are augmented with reversed triples in order to mimic the scoring of head entities as well. Plus, under the ZSLP setting, predictions on test triples are made w.r.t. a relation-dependent set of candidate tails $\mathcal{C}_{(h,r_{u})}$. This means that for a given relation, only semantically compatible tails are scored and ranked for reporting the final performance of the LP model.

It is worth mentioning that in both FSLP and ZSLP, for all testing triples $(h, r_{u}, t)$, $h, t \in \mathcal{E}$. In other words, each entity appearing in the testing set has already been observed in the training set. Performing predictions on both unseen entities and relations is a challenging setting barely explored in recent works~\cite{geng2021}.

Despite all the commonalities between FSLP and ZSLP, ZSLP differs from FSLP in that no training data is available on the relations in the test set. In other words, there is only a query set $\mathcal{Q}_r$ and no support set $\mathcal{S}_r$.

\begin{table}[htbp]
\centering
\footnotesize
\caption{Presentation of ZSLP works. The last three column headers refer to the incorporation of external knowledge, the nature of the unseen element (entity and/or relation), and the nature of the element to predict}\label{tab:zslp-models}
\begin{tabular}{p{2.25cm}p{0.5cm}p{1.5cm}p{2cm}p{0.5cm}p{0.25cm}}
\hline
Authors & Year & Model & Ext. Knowl. & U & P \\
\hline
Qin \textit{et al.} \cite{qin2020} & 2020 & ZSGAN & Text & R & E \\
Geng \textit{et al.} \cite{geng2021} & 2021 & OntoZSL & Ontology, Text & R & E \\
Wang \textit{et al.} \cite{wang2021ilpstar} & 2021 & StAR & Text & E, R & E \\
Zhang \textit{et al.} \cite{zhang2021fslzsl} & 2021 & GRL & \dashed & R & E \\
Chen \textit{et al.} \cite{chen2022zsltext} & 2022 & KG-S2S & Text & R & E \\
Cornell \textit{et al.} \cite{cornell2022} & 2022 & \dashed & Text & R & E \\
Geng \textit{et al.} \cite{geng2022} & 2022 & DOZSL & Ontology & R & E \\
Li \textit{et al.} \cite{li2022HAPZSL} & 2022 & HAPZSL & Text & R & E \\
Li \textit{et al.} \cite{li2022HNZSLP} & 2022 & HNZSLP & Text & R & E \\
Song \textit{et al.} \cite{song2022zsl} & 2022 & DMoG & Ontology, Text & R & E \\
Li \textit{et al.} \cite{li2023zslstructure} & 2023 & SEGAN & Text & R & E \\
Yu \textit{et al.} \cite{yu2023zsl} & 2023 & MASZSL & Text & R & E \\
\hline
\end{tabular}
\end{table}

\begin{table}[h]
\caption{Statistics of datasets used in ZSLP. Column headers from left to right: number of entities, triples, tasks (in train/dev/test splits), original source of the datasets, and references to those papers that use them. Dashed elements denote missing information}\label{tab:zslp-datasets}
	\centering\footnotesize
	\begin{tabular}{lccccc}
		\toprule
		Dataset  & \#Ent. & \#Triples & \#Tasks & Source & References\\
		\midrule
        DB100K-ZS & \dashed  & \dashed & \dashed & \cite{song2022zsl} & \cite{song2022zsl} \\
        FB15k/FB15k-237\footnote{FB15k-237 is used as the training dataset, and FB15k is used to sample $500$ triples whose relations are unseen during training.} & 14,541 & 281,624 & \dashed & \cite{zhang2021fslzsl} & \cite{zhang2021fslzsl} \\
        $\text{NELL}^{\heartsuit}$ & 68,544  & 192,797 & 342/5/11 & \cite{wang2021ilpstar} &  \cite{wang2021ilpstar, chen2022zsltext} \\
		NELL-ZS & 65,567  & 188,392 & 139/10/32 & \cite{qin2020} & \cite{qin2020, geng2021, geng2022, cornell2022, li2022HAPZSL, li2022HNZSLP, li2023zslstructure}\\
		Wiki-ZS & 605,812 & 724,967 & 469/20/48 & \cite{qin2020} & \cite{qin2020, geng2021, geng2022, cornell2022, li2022HAPZSL, li2022HNZSLP, li2023zslstructure}\\
		\bottomrule
	\end{tabular}
\end{table}

\subsubsection{Model Presentations.} Due to the absence of support triples for performing ZSLP, most works leverage auxiliary sources of information to extrapolate knowledge 
In overall, this information is of two types: ontologies, and textual descriptions of graph elements. ZSGAN~\cite{qin2020} relies on the latter type of information and uses a Generative Adversarial Network (GAN) to output high-quality relation embeddings for unseen relations. OntoZSL~\cite{geng2021} also aims at generating plausible embeddings for unseen relations using a GAN. To that end, it combines ontology embeddings for classes and relations with a text-aware encoder.
Following their work on OntoZSL, Geng \textit{et al.} propose DOZSL~\cite{geng2022}, a model that disentangles the embeddings of ontological concepts in order to obtain finer-grained inter-class relationships between seen and unseen relations. It uses a GAN-based generative model and a GCN-based propagation model to integrate disentangled embeddings for unseen relations.

Li \textit{et al.}~\cite{li2022HAPZSL} and Yu \textit{et al.}~\cite{yu2023zsl}  point out the issue of model collapse and training stability in former works. HAPZSL~\cite{li2022HAPZSL} addresses this by encoding relation prototype and the entity pair representation using an hybrid attention mechanism. MASZSL~\cite{yu2023zsl} establishes a connection between the structured KG semantic space and the unstructured text semantic space of emerging entities, forcing the entity pair to lie in a close neighborhood of their shared relations.

Li \textit{et al.}~\cite{li2022HNZSLP} note that proposed approaches so far use textual features of relations, which binds them to a fixed vocabulary and cannot handle out-of-vocabulary tokens. Their model HNZSLP leverages character n-gram information for ZSLP, thereby alleviating the aforementioned caveat.
GRL is a framework proposed in~\cite{zhang2021fslzsl}, which uses semantic correlations among relations as a bridge to connect semantically similar relations. Interestingly, GRL can be plugged in any transductive KGEM, such as TransE and ConvE.

Song \textit{et al.}~\cite{song2022zsl} propose to use three graphs: a factual graph, an ontology graph, and a textual graph. Their approach DMoG represents unseen relations in the factual graph by merging ontology with textual graphs, which are decoupled from the other graphs to avoid overfitting on seen relations.

StAR~\cite{wang2021ilpstar} partitions each triple into two asymmetric parts, both of which are encoded into contextualized representations by a Siamese-style textual encoder.
In~\cite{chen2022zsltext}, Chen \textit{et al.} propose KG-S2S, a model that converts different kinds of KG structures into a text-to-text format, thereby producing the text of target predicted entities.

Li \textit{et al.}~\cite{li2023zslstructure} affirm that ZSLP approaches relying on GANs ignores the gap between relation text description and relation structured representation. To bridge this gap, they propose SEGAN, which brings performance gains over GAN-based 
current methods.

\subsection{Inductive Link Prediction}\label{sec:ilp}
\subsubsection{Task Definition.} Compared to the FSL and ZSL settings that originate from other application fields such as computer vision, ILP is a task specifically designed for LP -- as the name suggests. ILP gained momentum recently. 

According to Galkin \textit{et al.}~\cite{galkin2022}, ILP goes beyond the transductive setting as it consists in training a model on one graph and moving to a new graph for running inference.
The set of relations of this \emph{inference} graph can either be the same as the training graph, or only a subset~\cite{ali2021}. Therefore, under this definition of the ILP setting, the inference graph does not contain unseen relations. 

Several works refine this general definition by subdividing it further. Namely, it is common to distinguish between \emph{fully-inductive} LP setting -- when the set of entities in the inference graph is disjoint from the set of entities seen during training -- and the \emph{semi-inductive} LP -- when the inference graph also contains observed entities~\cite{ali2021, galkin2022}.
The two situations -- when the inference graph contains only unseen entities vs. when it also contains entities observed during training -- have distinct consequences on the connectivity of the inference graph. In the former case, the inference graph is not connected to the training graph as their respective sets of entities are disjoint.
In the latter case, the inference graph is connected to the training graph.

However, other nomenclatures and taxonomies exist, which we will discuss in Section~\ref{sec:diverging-def}.

\begin{table}[htbp]
\centering
\scriptsize
\caption{Presentation of ILP works. The last four column headers refer to the incorporation of external knowledge, the learning paradigm, the nature of the unseen element (entity and/or relation), and the nature of the element to predict}\label{tab:ilp-models}
\begin{tabular}{p{2.5cm}p{0.5cm}p{1.5cm}p{2cm}p{4cm}p{0.45cm}p{0.45cm}}
\hline
Authors & Year & Model & Ext. Knowl. & Paradigm & U & P \\
\hline
Zhao \textit{et al.}\footnote{Although the authors identify their work as ZSLP, their experimental procedure corresponds to ILP as defined in Section~\ref{sec:ilp}.} \cite{zhao2017} & 2017 & JointE & Text & Entity encoding & E & E \\
Hamaguchi \textit{et al.} \cite{hamaguchi2017} & 2017 & MEAN & \dashed & Entity encoding & E & E \\
Shi \textit{et al.} \cite{shi2018ilp} & 2018 & ConMask & Text & Entity encoding & E & E \\
Tagawa \textit{et al.} \cite{tagawa2019} & 2019 & \dashed & Text & Entity encoding & E & E \\
Shah \textit{et al.} \cite{shah2019ilp} & 2019 & OWE & Text & Entity encoding & E & E \\
Wang \textit{et al.} \cite{wang2019LAN} & 2020 & LAN & \dashed & Entity encoding & E & E \\
Albooyeh \textit{et al.} \cite{albooyeh2020} & 2020 & OOS & \dashed & Entity encoding & E & E \\
Baek \textit{et al.} \cite{baek2020} & 2020 & GEN & \dashed & Entity encoding & E & E \\
Bi \textit{et al.} \cite{bi2020} & 2020 & \dashed & \dashed & Entity encoding & E & E \\
Bhowmik \textit{et al.} \cite{bhowmik2020} & 2020 & ELPE & \dashed & Entity encoding & E & E \\
He \textit{et al.} \cite{he2020ilp} & 2020 & VN Network & \dashed & Entity encoding & E & E \\
Shah \textit{et al.} \cite{shah2020ilp} & 2020 & \dashed & Text & Entity encoding & E & E \\
Zhang \textit{et al.} \cite{zhang2020OOV} & 2020 & GCN+AFM & Text & Entity encoding & E & E \\
Clouatre \textit{et al.} \cite{clouatre2021} & 2021 & MLMLM & Text & Entity encoding & E & E \\
Dai \textit{et al.} \cite{dai2021ilp} & 2021 & InvTransE & \dashed & Entity encoding & E & E \\
Daza \textit{et al.} \cite{daza2021ilp} & 2021 & BLP & Text & Entity encoding & E & E \\
Demir \textit{et al.} \cite{demir2021} & 2021 & \dashed & \dashed & Entity encoding & E & E \\
Liu \textit{et al.} \cite{liu2021ilp} & 2021 & INDIGO & \dashed & Entity encoding & E & E \\
Ren \textit{et al.} \cite{ren2021ilp} & 2021 & CatE & Ontology & Entity encoding & E & E \\
Wang \textit{et al.} \cite{wang2021ilpKEPLER} & 2021 & KEPLER & Text & Entity encoding & E & E \\
Wang \textit{et al.} \cite{wang2021ilpstar} & 2021 & StAR & Text & Entity encoding & E, R & E \\
Zhang \textit{et al.} \cite{zhang2021OOKB} & 2021 & HRFN & \dashed & Entity encoding & E & E \\
Chen \textit{et al.} \cite{chen2022morseilp} & 2022 & MorsE & \dashed & Entity encoding & E & E \\
Cui \textit{et al.} \cite{cui2022ilp} & 2022 & ARGCN & \dashed & Entity encoding & E & E \\
Galkin \textit{et al.} \cite{galkin2022ilp} & 2020 & NodePiece & \dashed & Entity encoding & E & E \\
Gesese \textit{et al.} \cite{gesese2022} & 2022 & RAILD & Text & Entity encoding & E, R & E \\
Li \textit{et al.} \cite{li2022SLAN} & 2022 & SLAN & \dashed & Entity encoding & E & E \\
Markowitz \textit{et al.} \cite{markowitz2022ilp} & 2022 & StATIK & Text & Entity encoding & E & E \\
Oh \textit{et al.} \cite{oh2022} & 2022 & IKGE & Text & Entity encoding & E, R & E, R \\
Wang \textit{et al.} \cite{wang2022ilp} & 2022 & CFAG & \dashed & Entity encoding & E & E \\
Wang \textit{et al.} \cite{wang2022ilpsimkgc} & 2022 & SimKGC & Text & Entity encoding & E & E \\
Zha \textit{et al.} \cite{zha2022ilpbert} & 2022 & BertRL & \dashed & Entity encoding & E & E \\
Lee \textit{et al.} \cite{lee2023ilp} & 2023 & InGram & \dashed & Entity encoding & E, R & E \\
Li \textit{et al.} \cite{li2023ilp} & 2023 & REPORT & \dashed & Entity encoding & E & E \\
Li \textit{et al.} \cite{li2023ilpcontrastive} & 2023 & VMCL & \dashed & Entity encoding & E & E \\
Samy \textit{et al.} \cite{samy2023ilp} & 2023 & Graph2Feat & \dashed & Entity encoding & E & E \\
Zhao \textit{et al.} \cite{zhao2023fslpilp} & 2023 & RawNP & \dashed & Entity encoding & E & E \\
Wang \textit{et al.} \cite{wang2021ilppathcon} & 2021 & PathCon & \dashed & Subgraph enclosing & E & E \\
Teru \textit{et al.} \cite{teru2020} & 2020 & GraIL & \dashed & Subgraph enclosing & E & E \\
Chen \textit{et al.} \cite{chen2021TACT} & 2021 & TACT & \dashed & Subgraph enclosing & E & E \\
Mai \textit{et al.} \cite{mai2021ilp} & 2021 & COMPILE & \dashed & Subgraph enclosing & E & E \\
Zhu \textit{et al.} \cite{zhu2021ilp} & 2021 & NBFNet & \dashed & Subgraph enclosing & E & E \\
Kwak \textit{et al.} \cite{kwak2022ilp} & 2022 & SGI & \dashed & Subgraph enclosing & E & E \\
Jin \textit{et al.} \cite{jin2022ilp} & 2022 & GraphANGEL & \dashed & Subgraph enclosing & R & E \\
Lin \textit{et al.} \cite{lin2022ilp} & 2022 & ConGLR & \dashed & Subgraph enclosing, Rule-based & E & E \\
Pan \textit{et al.} \cite{pan2022ilp} & 2022 & LogCo & \dashed & Subgraph enclosing, Rule-based & E & E \\
Xu \textit{et al.} \cite{xu2022ilp} & 2022 & SNRI & \dashed & Subgraph enclosing & E, R & E \\
Yan \textit{et al.} \cite{yan2022ilp} & 2022 & CBGNN & \dashed & Subgraph enclosing, Rule-based & E & E \\
Yang \textit{et al.} \cite{yang2022fslpilp} & 2022 & MILP & \dashed & Subgraph enclosing & E & E \\
Zhang \textit{et al.} \cite{zhang2022ilp} & 2022 & RED-GNN & \dashed &Subgraph enclosing  & E & E \\
Geng \textit{et al.} \cite{geng2023} & 2023 & RMPI & Ontology & Subgraph enclosing & E, R & E \\
Mohamed \textit{et al.} \cite{mohamed2023ilp} & 2023 & LCILP & \dashed & Subgraph enclosing & E & E \\
Su \textit{et al.} \cite{su2023ilp} & 2023 & KRST & Text & Subgraph enclosing & E & E \\
Xie \textit{et al.} \cite{xie2023ilpanchor} & 2023 & QAAR & \dashed & Subgraph enclosing & E & E \\
Zhang \textit{et al.} \cite{zhang2023DEKG} & 2023 & DEKG-ILP & \dashed & Subgraph enclosing & E, R & E, R \\
Zheng \textit{et al.} \cite{zheng2023fslpilp} & 2023 & Meta-iKG & \dashed & Subgraph enclosing & E & E \\
Sadeghian \textit{et al.} \cite{sadeghian2019} & 2019 & DRUM & \dashed & Rule-based & E & E \\
\hline
\end{tabular}
\end{table}

\begin{table}[htbp]
\centering\footnotesize
\caption{Statistics of datasets used in ILP. Horizontal lines separate datasets based on the underlying knowledge graph: Freebase, NELL, WordNet, DBPedia, YAGO, and Wikidata, respectively}\label{tab:ilp-datasets}
\begin{tabular}{lll}
\toprule
Dataset & Source & References \\
\midrule
FB15k-237-v{1/2/3/4} & \cite{teru2020} & \cite{pan2022ilp, lin2022ilp, mai2021ilp, geng2023, zhang2021OOKB, xu2022ilp, kwak2022ilp, li2023ilpcontrastive, xie2023ilpanchor, mohamed2023ilp, chen2021TACT, teru2020, zhang2022ilp, zhu2021ilp, liu2021ilp, li2023ilp, zhu2021ilp} \\
$\text{FB15k-237}^{\dagger}$ & \cite{daza2021ilp} & \cite{daza2021ilp, markowitz2022ilp, gesese2022} \\
$\text{FB15k-237}^{\heartsuit}$ & \cite{baek2020} & \cite{baek2020, zhang2021OOKB} \\
FB15k-Head-{5/10/15/20/25} & \cite{wang2019LAN} & \cite{wang2019LAN, dai2021ilp, li2022SLAN} \\
FB15kTail-{5/10/15/20/25} & \cite{wang2019LAN} & \cite{wang2019LAN, dai2021ilp, li2022SLAN} \\
FB-{25/50/75/100} & \cite{lee2023ilp} & \cite{lee2023ilp} \\
FB-MBE & \cite{cui2022ilp} & \cite{cui2022ilp} \\
FB-{EQ/MB/ME} & \cite{zhang2023DEKG} & \cite{zhang2023DEKG} \\
FB15k-237-{1K/2K/Full} & \cite{zha2022ilpbert} & \cite{zha2022ilpbert} \\
FB15k-237(10/20\%) & \cite{li2022SLAN} & \cite{li2022SLAN} \\
FB15k-237-Inductive & \cite{bhowmik2020} & \cite{bhowmik2020} \\
oFB15k-237 & \cite{albooyeh2020} & \cite{albooyeh2020} \\
OW\_FB15K-237 & \cite{zhang2020OOV} & \cite{zhang2020OOV} \\
FB15k-237-OWE & \cite{shah2019ilp} & \cite{shah2019ilp, shah2020ilp} \\
FB13-{Head/Tail/Both}-{1K/3K/5K} & \cite{hamaguchi2017} & \cite{hamaguchi2017} \\
FB20k & \cite{xie2016ilp} & \cite{xie2016ilp, tagawa2019, shah2019ilp, shah2020ilp} \\
FB20k+ & \cite{oh2022} & \cite{oh2022} \\
FB-Ext & \cite{chen2022ilpIJCAI} & \cite{chen2022ilpIJCAI} \\
\midrule
NELL-995-v{1/2/3/4} & \cite{teru2020} & \cite{pan2022ilp, lin2022ilp, mai2021ilp, geng2023, zhang2021OOKB, kwak2022ilp, li2023ilpcontrastive, xie2023ilpanchor, mohamed2023ilp, chen2021TACT, teru2020, zhang2022ilp, zhu2021ilp, liu2021ilp} \\
NELL-995 & \cite{baek2020} & \cite{baek2020, zhang2021OOKB} \\
NELL-995-Inductive & \cite{bhowmik2020} & \cite{bhowmik2020} \\
NELL-MBE & \cite{cui2022ilp} & \cite{cui2022ilp} \\
NELL-{EQ/MB/ME} & \cite{zhang2023DEKG} & \cite{zhang2023DEKG} \\
NELL-995-{1K/2K/Full} & \cite{zha2022ilpbert} & \cite{zha2022ilpbert, su2023ilp} \\
NELL-Ext & \cite{chen2022ilpIJCAI} & \cite{chen2022ilpIJCAI} \\
NL-{25/50/75/100} & \cite{lee2023ilp} & \cite{lee2023ilp} \\
\midrule
WN18RR-v{1/2/3/4} & \cite{teru2020} & \cite{pan2022ilp, lin2022ilp, mai2021ilp, geng2023, xu2022ilp, xie2023ilpanchor, mohamed2023ilp, chen2021TACT, teru2020, zhang2022ilp, zhu2021ilp, liu2021ilp, li2023ilp, zhu2021ilp} \\
$\text{WN18RR}^{\dagger}$ & \cite{daza2021ilp} & \cite{daza2021ilp, markowitz2022ilp, gesese2022} \\
$\text{WN18RR}^{\heartsuit}$ & \cite{clouatre2021} & \cite{clouatre2021} \\
WN18RR-Inductive & \cite{bhowmik2020} & \cite{bhowmik2020} \\
WN-MBE & \cite{cui2022ilp} & \cite{cui2022ilp} \\
WN-{EQ/MB/ME} & \cite{zhang2023DEKG} & \cite{zhang2023DEKG} \\
WN18RR-{1K/2K/Full} & \cite{zha2022ilpbert} & \cite{zha2022ilpbert, su2023ilp} \\
WN18RR(10/20\%) & \cite{li2022SLAN} & \cite{li2022SLAN} \\
oWN18RR & \cite{albooyeh2020} & \cite{albooyeh2020} \\
WN11-{Head/Tail/Both}-{1K/3K/5K} & \cite{hamaguchi2017} & \cite{hamaguchi2017, bi2020} \\
\midrule
DBPedia50k & \cite{shi2018ilp} & \cite{shi2018ilp, shah2019ilp, shah2020ilp, zhang2020OOV} \\
DBPedia50k+ & \cite{oh2022} & \cite{oh2022} \\
DBPedia500k & \cite{shi2018ilp} & \cite{shi2018ilp} \\
DBPedia500k+ & \cite{oh2022} & \cite{oh2022} \\
DB111k-174 & \cite{ren2021ilp} & \cite{ren2021ilp} \\
\midrule
YAGO26k-906 & \cite{ren2021ilp} & \cite{ren2021ilp} \\
YAGO3-10(10/20\%) & \cite{li2022SLAN} & \cite{li2022SLAN} \\
Yago37k-{5/10/15/20/25} & \cite{he2020ilp} & \cite{he2020ilp} \\
\midrule
Wikidata5M & \cite{wang2021ilpKEPLER} & \cite{wang2021ilpKEPLER, daza2021ilp, clouatre2021, markowitz2022ilp, wang2022ilpsimkgc, gesese2022} \\
WK-{25/50/75/100} & \cite{lee2023ilp} & \cite{lee2023ilp} \\
ILPC22-S & \cite{galkin2022} & \cite{galkin2022} \\
ILPC22-L & \cite{galkin2022} & \cite{galkin2022} \\
\bottomrule
\end{tabular}
\end{table}

\subsubsection{Model Presentations.} While Demir \textit{et al.}~\cite{demir2021} investigated the impact of unseen entities in the performance of transductive KGEMs, special approaches are usually needed for handling the ILP task. These approaches are broadly divided into two main families: entity encoding, and subgraph enclosing models. Entity encoding models transfer knowledge from seen to unseen entities by aggregating neighbors of the unseen entities. In contrast, subgraph enclosing models start by extracting the relational subgraph between the head and tail of a triple and encode them as a pair. The underlying heuristics is that the subgraph between these two entities conveys much of the semantics that link them, which can be leveraged for prediction.

\sstitle{Entity encoding} models are represented by the work from Hamaguchi \textit{et al.}~\cite{hamaguchi2017}, who propose MEAN as one of the earliest GNN-based model for propagating entity representations and embed unseen emerging entities with different transition and pooling functions. A more efficient entity neighbor aggregator can be found in LAN~\cite{wang2019LAN}, which derives attention weights based on logic rules. VN Network~\cite{he2020ilp} is an even more sophisticated model that relies on symmetric path rules.

GNNs underpin many recent works targeting the ILP task, due to their inherent inductive capabilities~\cite{liu2021ilp}. INDIGO~\cite{liu2021ilp} is a GNN-based model that encodes KGs using a one-to-one correspondence between triples and elements of node feature vectors in the graphs processed by the GNN.
NodePiece~\cite{galkin2022ilp} leverages inductive node and relational structure representations and is able to conduct logical reasoning at inference time on very large KGs. Graph2Feat~\cite{samy2023ilp} is a recently proposed GNN-based model using knowledge distillation, combining a lightweight student multi-layer perceptron (MLP) with a more expressive teacher GNN. Therefore, Graph2Feat combines the expressiveness of GNNs with the fast inference capabilities of MLPs.
IKGE~\cite{oh2022} leverages an attentive feature aggregation mechanism to hierarchically accumulate the features of multi-hop neighbors and generate more expressive embeddings for unseen entities.
Bi \textit{et al.}~\cite{bi2020} propose a novel and parameter-efficient model that draws from the advantages of both GNN and CNN-based methods.

Another family of approaches extends transductive models. Albooyeh \textit{et al.}~\cite{albooyeh2020} propose modified versions of DistMult, namely oDistMult-ERAvg and oDistMult-LS. Dai \textit{et al.}~\cite{dai2021ilp} present a simple yet effective method to inductively represent unseen entities by their optimal estimation under translational assumptions. Building on TransE and RotatE models, they consequently propose Inv-TransE and Inv-RotatE.

Bhowmik \textit{et al.}~\cite{bhowmik2020} propose one of few methods that provide explainable reasoning paths for the predictions made on unseen entities.

CFAG~\cite{wang2022ilp} utilizes the
relational semantics at two granularity levels to better represent emerging entities and improve predictive performance.

A novel perspective on the ILP task is to consider that unseen entities appear in batches. Cui \textit{et al.} propose ARGCN~\cite{cui2022ilp}, a model based on the GCN architecture that aims at tackling the multi-batch emergence scenario.
Also based on GCNs, Zhang \textit{et al.}~\cite{zhang2021OOKB} propose HRFN, a two-stage GCN-based model that first learns pre-representations for emerging entities using hyper-relation features learnt from the training set. Then, they use a feature aggregation strategy involving an entity-centered GCN and a relation-centered GCN.

Based on the assumption that similar entities usually appear in common graph contexts, SLAN~\cite{li2022SLAN} is a model built on a similarity-aware aggregation network. Distance between each entity pair is measured using a novel similarity-aware function that considers the contextual gap between such pairs of entities.

VMCL~\cite{li2023ilpcontrastive} is a model based on contrastive learning and VAEs. VMCL first uses representation generation to capture the generated representations of entities. Then, the generated variations can augment the representation space with complementary features.

As existing neighbors of emerging entities may be too sparse to provide enough information, CatE~\cite{ren2021ilp} solves this sparsity issue with the enhancement from ontological concepts.

InGram~\cite{lee2023ilp} is a recent model that can generate embeddings for both unseen entities and relations. It first generates a relation graph as a weighted graph containing relations and their affinity weights. It subsequently uses this relation graph with the original KG to generate the representations of entities and relations using an attention mechanism. RAILD~\cite{gesese2022} is also based on the idea of creating an intermediary graph for the purpose of generating relation representations. It uses a relation-relation network from the contextual information present in the graph structure, which helps producing better embeddings for unseen relations. MaKEr~\cite{chen2022federated} is also able to generate representations for both unseen entities and relations by extending the ILP task to the federated setting, and applying a meta-learning-based approach to mimic the ILP task in the test KG. It is able to generalize to unseen components of the test KG after being meta-trained. Although only targeting unseen entities, MorsE~\cite{chen2022morseilp} also resorts to transferable meta-knowledge to produce entity embeddings. Such meta-knowledge is modeled by entity-independent modules and learned by meta-learning.

Various entity-encoding-based models leverage additional textual information. The very first work fusing entity descriptions to produce structure-based and description-based embeddings jointly is DKRL~\cite{xie2016ilp}.
Tagawa \textit{et al.}~\cite{tagawa2019} propose a method to learn entity representations via a graph structure that uses both seen and unseen entities as well as words. The latter are considered as nodes created from the descriptions of all entities. 
Similarly, ConMask~\cite{shi2018ilp} implements relationship-dependent content masking on entity descriptions and uses the extracted description-based representations to make predictions.
OWE~\cite{shah2019ilp} adopts a different approach to the ILP problem by training traditional KGEMs and word embeddings independently,
and learns a transformation from the textual description of entities
to their embeddings.
Shah \textit{et al.}~\cite{shah2020ilp} propose learning a relation-specific transformation functions from a text-based embedding space to a graph-based embedding space, where a transductive LP model can be applied.
GCN+AFM~\cite{zhang2020OOV} is built on a pairwise factorization model and exploits a neural bag-of-word-based encoder, which encodes word-based entities into entity embeddings for the decoder.

A broad range of recent works fusing textual descriptions with structure-based information leverages Language Models (LMs). BLP~\cite{daza2021ilp} uses a pre-trained LM as an entity encoder and fine-tunes it for ILP.
KEPLER~\cite{wang2021ilpKEPLER} also uses a pre-trained LM to encode entities and texts. These are jointly optimized using a masked language modeling objective.
Similarly, MLMLM~\cite{clouatre2021} is a model for training masked language models to perform ILP on unseen entities. 
StAR~\cite{wang2021ilpstar} is a structure-augmented text representation model where triples are encoded via a Siamese-style encoder and where the scoring module combines two parallel scoring strategies to learn both contextualized and structured knowledge. 
SimKGC~\cite{wang2022ilpsimkgc} adopts a bi-encoder architecture and combines three types of negative samples.
StATIK~\cite{markowitz2022ilp} combines a LM with a message-passing-based GNN to build a hybrid encoder, and uses TransE as the decoder for scoring triples.
BertRL~\cite{zha2022ilpbert} fine-tunes a pre-trained LM by taking relation instances and their possible reasoning paths as training samples. 
REPORT~\cite{li2023ilp} is a recent model that fully uses relational paths and context, and aggregates them adaptively within a unified hierarchical Transformer architecture. In the same vein as what Bhowmik \textit{et al.}~\cite{bhowmik2020} propose, REPORT can provide explanations for each prediction it makes.

\sstitle{Subraph encoding} models are also often used for the ILP task. One of the earliest representative models is GraIL~\cite{teru2020}. GraIL is a GNN-based model that starts by extracting the enclosing subgraph between two unseen entities. Then, each entity present in the extracted subgraph is labeled based on its distance with respect to the two unseen entities. 
TACT~\cite{chen2021TACT} improves over GraIL by additionally considering inter-relation correlations in the extracted subgraphs.
COMPILE~\cite{mai2021ilp} is based on a node-edge communicative message-passing mechanism, which also helps considering the relative importance of different relations. COMPILE is reported to handle asymmetric relations with more accuracy than current approaches.

The path between two entities can also be viewed as a subgraph. PathCon~\cite{wang2021ilppathcon} combines two types of subgraph structures: the contextual relations as well as the relational paths between head and tail entity. 
NBFNet~\cite{zhu2021ilp} parameterizes the generalized Bellman-Ford algorithm with learned indicator, message, and aggregator functions to enhance the overall expressiveness and inductive capabilities of GNNs.  
RED-GNN~\cite{zhang2022ilp} introduces the \emph{r-digraph} as a novel structure of relational paths for LP. Inspired by solving overlapping sub-problems by dynamic programming, RED-GNN builds the r-digraph to improve over baselines regarding the ILP task setting. GraphANGEL~\cite{jin2022ilp} introduces several graph pattern searching and sampling techniques, which can efficiently find subgraphs matching the patterns in triangle and quadrangle shapes. KRST~\cite{su2023ilp} is designed to encode the extracted reliable paths in KGs, allowing to cluster paths and provide multi-aspect explanations for each prediction.

By regarding the relational path and the target relation as the body and head of a logic rule, respectively, LogCo~\cite{pan2022ilp} allows to perform ILP by fusing GNNs with discrete logic and relational paths. Also combining logic rules with GNNs, CBGNN~\cite{yan2022ilp} tackles the ILP task through a cycle-centric perspective.
CBGNN runs implicit algebraic operations in the cycle space
through the message passing of a GNN in order to learn the
representation of good rules.

ConGLR~\cite{lin2022ilp} incorporates context graph with logical reasoning. Two GCNs are fed with the information interaction of entities and relations and are responsible for processing the subgraph and context graph, respectively.
Based on the observation that most works only consider the enclosing part of subgraph without complete neighboring relations, SNRI~\cite{xu2022ilp} is further proposed to handle sparse subgraphs by leveraging the complete neighbor
relations of entities using extracted neighboring relational
features and paths.

Compared with previous GCN-based methods, such as GraIL, CoMPILE, and ConGLR, QAAR~\cite{xie2023ilpanchor} is able to capture richer transferable entity-independent features to perform ILP on subgraphs.

LCILP~\cite{mohamed2023ilp} proposes a new way of performing subgraph extraction, based on a local clustering procedure and a personalized PageRank approach.

Whereas generating high-quality negative samples in transductive LP is a vibrant research area, Kwak \textit{et al.}~\cite{kwak2022ilp} point out that there is currently no method for selecting hard negatives for the ILP task. Therefore, they propose SGI to overcome this limitation. SGI relies on a novel sampling method for selecting hard negative samples and a newly-proposed training objective which maximizes the mutual information between the target relation and the enclosing subgraph.

Few models are actually proposed to handle both unseen entities and relations. RMPI~\cite{geng2023} tackles this issue by leveraging the relation semantics defined in ontological schemas. DEKG-ILP~\cite{zhang2023DEKG} is also concerned with this more challenging setting. It features a contrastive learning-based and relation-specific module to extract global relation-based semantic features that are shared between original KGs and disconnected emerging KGs with a newly-proposed sampling strategy.

\sstitle{Rule-based} models are intrinsically inductive, as they do not depend on any specific entities.
Neural LP~\cite{neurallp2017} is based on TensorLog and implements an end-to-end differentiable method for learning the parameters as well as the structure of logical rules. DRUM~\cite{sadeghian2019} is a scalable and differentiable model for mining first-order logical rules from KGs. DRUM uses bidirectional RNNs to share useful information across relation-specific tasks.
RLogic~\cite{rlogic2022} takes a different perspective on the problem and defines a predicate representation learning-based scoring model trained by sampled rule instances.

\section{Beyond the Transductive Setting: the Current State of Affairs}\label{sec:current-state}
In this section, we raise concerns about the ILP, FSLP, and ZSLP tasks as introduced in the previous section. In particular, we highlight the need for agreeing on a common suite of benchmark datasets in Section~\ref{sec:benchmarks}, due to the significant disparity in the datasets used in practice. In addition, we point out the undesirable existence of many different terminologies regarding the ILP task. In particular, different authors use distinct naming conventions to conceptually describe the same task or, conversely, use the same terminology to refer to intrinsically different practical setups (Section~\ref{sec:diverging-def}).
We reflect on the actual definitions of ILP, FSLP, and ZSLP in Section~\ref{sec:towards} and ultimately propose a comprehensive and unifying framework for categorizing these works in Section~\ref{sec:nomenclature}.  

\subsection{A lack of generalized benchmarks}\label{sec:benchmarks}
\begin{figure}[htbp]
  \centering
  \includegraphics[width=0.75\textwidth]{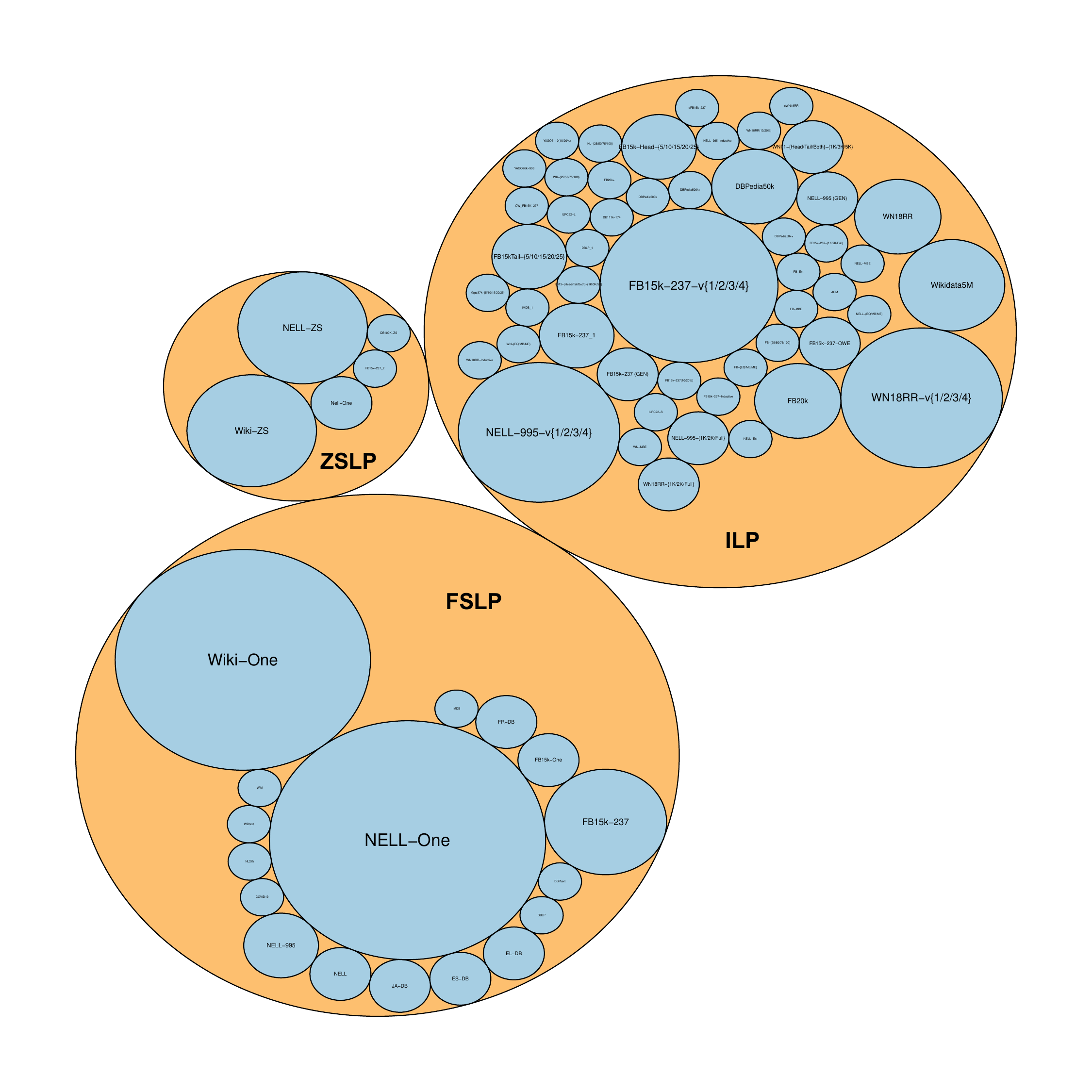}
  \caption{Circular packing graph representing dataset usage in ILP, FSLP, and ZSLP settings}
  \label{fig:circular-stacking}
\end{figure}

Fig.~\ref{fig:circular-stacking} depicts the datasets used under each setting -- ILP, FSLP, and ZSLP. The three outer circles represent ILP, FSLP, and ZSLP. The inner ones represent the datasets used under the corresponding setting. The size of the outer circles is proportional to the number of papers published for a given setting, while the size of the inner circles denote the number of papers using a given dataset, \textit{i.e.} the prevalence of datasets. Note that the size of these inner circles is \emph{only} proportional to the usage of datasets under a given setting. Therefore, Fig.~\ref{fig:circular-stacking} does not allow to deduce whether a dataset is systematically used in (almost) all papers referring to a particular setting. For this, we provide a complete description of dataset usage in Table~\ref{tab:fslp-datasets}, Table~\ref{tab:zslp-datasets}, and Table~\ref{tab:ilp-datasets}.

Notably, we observe fewer datasets used under the ZSLP setting. Obviously, this is because ZSLP is less explored than FSLP and ILP. However, NELL-ZS and Wiki-ZS are datasets used in almost all the retrieved ZSLP papers (Table~\ref{tab:zslp-datasets}) and are therefore recognized as public benchmarks.

The situation is less clear for FSLP: although NELL-One and Wiki-One are arguably well-recognized benchmarks, we observe that a handful of additional datasets are used in practice. It is worth noting that most of them are used in a single work or in a few works of the same authors (Table~\ref{tab:fslp-datasets}). We claim that using disparate datasets severely restricts the relevance of model comparisons. As for the FSLP task, more specifically, we additionally observe different choices of $K$ for the few-shot meta-tasks (Table~\ref{tab:fslp-datasets}). Similarly, such arbitrary choices raise natural concerns over the generalization capabilities of presented models and approaches.

Datasets used in works that identify themselves as ILP-related are even more heterogeneous. However, most of them derive from FB15K-237~\cite{toutanova2015}, NELL-995~\cite{xiong2017}, and WN18RR~\cite{conve}. Notably, Teru \textit{et al.} sample 4 different inductive versions of the aforementioned datasets for tackling the ILP task with unseen $\leftrightarrow$ unseen test entities. For each dataset, the different proposed versions comprise an increasing number of nodes and edges. Since then, these 12 datasets have been extensively used in many ILP-related works, as evidenced in Table~\ref{tab:ilp-datasets}. However, there is a huge variety in choices of data filtering, train/valid/test splits, and more.

We argue that the lack of widely accepted benchmarks -- primarily for ILP, but also for FSLP -- severely restricts any attempt at comparing models fairly.
    
\subsection{A terminological drift for the ILP task}\label{sec:diverging-def}

In Section~\ref{sec:ilp}, we stuck to the most common definition of ILP~\cite{ali2021,galkin2022} to introduce the setting, which is further divided into semi-inductive and fully-inductive LP. In the current section, we present and discuss other perspectives and naming conventions existing in the literature.

\subsubsection{Marginal Terminologies}
Table~\ref{tab:marginal-terminologies} reports various terms found in the selected papers from the existing literature. These terms do not aim at defining what inductive LP is \textit{per se}, but qualify the inductive aspect of predictions. 
This collection of supplementary terms to refer to the inductive part of LP raises the question of whether these terms convey any useful additional information compared to the term ``inductive''. If not, relying on a limited set of terms would surely eliminate any risk of confusion.

\begin{table}[t]
  \centering
  \caption{Marginal terminologies and corresponding papers}\label{tab:marginal-terminologies}
  \begin{tabular}{ll}
    \hline
    \textbf{Terminology} & \textbf{Papers}\\
    \hline
    Emerging entities & ~\cite{bhowmik2020, he2020ilp, zhang2023DEKG} \\
    Open-world & ~\cite{shi2018ilp, shah2019ilp, shah2020ilp} \\
    Out-of-distribution & ~\cite{zhou2022} \\
    Out-of-graph & ~\cite{baek2020} \\
    Out-of-knowledge-base & ~\cite{hamaguchi2017, bi2020, zhang2021OOKB} \\
    Out-of-knowledge-graph & ~\cite{dai2021ilp, oh2022, li2022SLAN} \\
    Out-of-sample & ~\cite{albooyeh2020} \\
    Out-of-vocabulary & ~\cite{zhang2020OOV, demir2021} \\
    Unseen entities & ~\cite{tagawa2019, clouatre2021} \\
    \hline
  \end{tabular}
\end{table}

\subsubsection{Different Definitions}
In Section~\ref{sec:ilp}, the ILP definition of Galkin \textit{et al.}~\cite{galkin2022} was presented. As we will see, this is an arbitrary choice, as many other definitions exist and convey a conceptualization of the predictive task which may differ in certain cases.

In particular, Gesese \textit{et al.}~\cite{gesese2022} subdivide ILP into three concrete settings: \emph{semi-inductive}, \emph{fully-inductive}, and \emph{truly-inductive} LP. Daza \textit{et al.}~\cite{daza2021ilp} as well as Markowitz \textit{et al.}~\cite{markowitz2022ilp} refer to \emph{dynamic} vs. \emph{transfer} LP settings. Zhang \textit{et al.}~\cite{zhang2023DEKG} distinguish between \emph{bridging}, \emph{enclosing}, and \emph{extended} inductive LP. As for Geng \textit{et al.}~\cite{geng2023}, they separate \emph{partially-inductive} LP from \emph{fully inductive} LP.

In what follows, we aim at presenting the different perspectives adopted in the aforementioned ILP-related works, highlighting thereby the existence of a significant terminological disparity when defining inductive LP.

Recall that Galkin \textit{et al.}~\cite{galkin2022} distinguish the \emph{fully-inductive} LP setting from the \emph{semi-inductive} LP one. The fully-inductive LP scenario corresponds to the case mentioned earlier where the inference graph is totally disconnected from the training graph, as it features triples with only unseen entities. The semi-inductive LP scenario corresponds to the case where the inference graph comprises a combination of seen and unseen entities. In both cases, the set of relations of the inference graph either equals or is a subset of the relation set observed on the training graph.

More formally, given the graph observed at training time $\mathcal{KG}_{train}$ and the inference graph at testing time $\mathcal{KG}_{inf}$ and with the common requirement that $\mathcal{R}(\mathcal{KG}_{inf}) \subseteq \mathcal{R}(\mathcal{KG}_{train})$:
\begin{description}
\item[\textbf{Semi-inductive setting}]  $\mathcal{E}(\mathcal{KG}_{inf}) \cap \mathcal{E}(\mathcal{KG}_{train}) \neq \emptyset$, \textit{i.e.} the inference graph is connected to the training graph through a shared subset of entities.
\item[\textbf{Fully-inductive setting}] $\mathcal{E}(\mathcal{KG}_{inf}) \cap \mathcal{E}(\mathcal{KG}_{train}) = \emptyset$, \textit{i.e.} the inference graph contains only unseen entities.
\end{description}

In view of these definitions provided in \cite{galkin2022}, in the fully-inductive setting it is unclear whether triples in the inference graph have to feature one seen entity and one unseen entity exactly, or if triples featuring two unseen entities are also allowed to exist. In other words, no detail is provided on whether unseen entities (either head or tail) in inference graph triples \emph{have} to be connected to entities observed in the training graph (the other position in the triple).

Gesese \textit{et al.}~\cite{gesese2022} address this with a triple-based view for defining their scenarios. In particular, they subdivide ILP into three settings: \emph{semi-inductive}, \emph{fully-inductive}, and \emph{truly-inductive}. Given a knowledge graph $\mathcal{KG} = \{ \mathcal{E}, \mathcal{R},  \mathcal{T}\}$, a training set $\mathcal{T}_{train}$ and a testing set $\mathcal{T}_{test}$ where $\mathcal{E}_{train}$ and $\mathcal{R}_{train}$, $\mathcal{E}_{test}$ and $\mathcal{R}_{test}$ are their corresponding set of entities and relations\footnote{The following definitions will be based on these assumptions as well. These are consequently omitted in the rest of the section.}, they define the three settings as follows:
\begin{description}
    \item[\textbf{Semi-inductive setting}] For every triple $(h,r,t) \in  \mathcal{T}_{test}$, either or both of $h \in \mathcal{E}_{train}$ and $t \in \mathcal{E}_{train}$ holds true while $\mathcal{R}_{test} \subseteq \mathcal{R}_{train}$.
    \item[\textbf{Fully-inductive setting}] For every triple $(h,r,t) \in \mathcal{T}_{test}$, both $h \notin \mathcal{E}_{train}$ and $t \notin \mathcal{E}_{train}$ holds true while $\mathcal{R}_{test} \subseteq \mathcal{R}_{train}$.
    \item[\textbf{Truly-inductive setting}] For every triple $(h,r,t) \in  \mathcal{T}_{test}$, either or both of $h \notin \mathcal{E}_{train}$ and $t \notin \mathcal{E}_{train}$ holds true while there exists a set $\mathcal{R'} \subseteq \mathcal{R}_{test}$ where $\mathcal{R'} \nsubseteq \mathcal{R}_{train}$.
\end{description}

First, Gesese \textit{et al.} present a new perspective on ILP-related settings compared to previous works. Specifically, while ILP has traditionally been approached with unseen entities and (a subset of) seen relations, their truly-inductive setting is meant to handle both unseen entities and a mix of seen and unseen relations. This directly questions the nature of the object to be predicted under the ILP scenario: should the inference graph be restrained to unseen entities, or could it include unseen relations as well? The latter option would undoubtedly result in a closer alignment between FSL, ZSL and ILP, as in the existing literature it has almost been taken for granted that FSL and ZSL are based on predicting unseen relations while ILP aims at predicting unseen entities.

Secondly, it is worth mentioning that although semi-inductive LP also appears in the taxonomy proposed by Gesese \textit{et al.}, it is not defined the same way as Galkin \textit{et al.} do. In particular, the semi-inductive LP setting of Gesese \textit{et al.} state that either the head or the tail of a test triple must be known, whereas in the definition of Galkin \textit{et al.}, it is possible that both the head and the tail are unseen.

Adopting a similar triple-based view in their task definitions, Daza \textit{et al.}~\cite{daza2021ilp} and Markowitz \textit{et al.}~\cite{markowitz2022ilp} differentiate between \emph{dynamic} and \emph{transfer} inductive LP, and add constraints on incorrect entities. They define their two settings as follows:
\begin{description}
    \item[\textbf{Dynamic setting}] For every triple $(h,r,t) \in \mathcal{T}_{test}$, either or both of $h \notin \mathcal{E}_{train}$ and $t \notin \mathcal{E}_{train}$ holds true while $\mathcal{R}_{test} \subseteq \mathcal{R}_{train}$. Incorrect entities are in $\mathcal{E}_{train} \cup \mathcal{E}_{test}$.
    \item[\textbf{Transfer setting}] For every triple $(h,r,t) \in \mathcal{T}_{test}$, both $h \notin \mathcal{E}_{train}$ and $t \notin \mathcal{E}_{train}$ holds true while $\mathcal{R}_{test} \subseteq \mathcal{R}_{train}$. Incorrect entities are in $\mathcal{E}_{test}$.
\end{description}

The dynamic setting corresponds to the case where each test triple contains at least one unseen entity while the transfer setting is when test triples feature two unseen entities. In other words, their dynamic (resp. transfer) setting maps to the semi-inductive (resp. fully-inductive) setting of Galkin \textit{et al.}.

Zhang \textit{et al.}~\cite{zhang2023DEKG} also adopt a triple-based criterion to partition ILP into \emph{enclosing}, \emph{bridging}, and \emph{extended} ILP. Their three settings are defined as follows:
\begin{description}
    \item[\textbf{Enclosing setting}] For every triple $(h,r,t) \in  \mathcal{T}_{test}$, both $h \notin \mathcal{E}_{train}$ and $t \notin \mathcal{E}_{train}$ hold true while $\mathcal{R}_{test} \subseteq \mathcal{R}_{train}$.
    \item[\textbf{Bridging setting}] For every triple $(h,r,t) \in  \mathcal{T}_{test}$, either $h \notin \mathcal{E}_{train}$ or $t \notin \mathcal{E}_{train}$ holds true while $\mathcal{R}_{test} \subseteq \mathcal{R}_{train}$.
    \item[\textbf{Extended setting}] For every triple $(h,r,t) \in  \mathcal{T}_{test}$, either or both of $h \notin \mathcal{E}_{train}$ and $t \notin \mathcal{E}_{train}$ holds true while $\mathcal{R}_{test} \subseteq \mathcal{R}_{train}$.
\end{description}

The enclosing ILP scenario is equivalent to the fully inductive setting by Gesese \textit{et al.}. The bridging ILP scenario allows exactly one of the two entities to be observed during training. The extended ILP scenario is a combination of the aforementioned two scenarios as it allows triples to contain either one or two unseen entities.

Geng \textit{et al.}~\cite{geng2023} also subdivides ILP into distinct scenarios. More specifically, they distinguish between partially-inductive LP and fully inductive LP:
\begin{description}
    \item[\textbf{Partially-inductive setting}] For every triple $(h,r,t) \in \mathcal{T}_{test}$, both $h \notin \mathcal{E}_{train}$ and $t \notin \mathcal{E}_{train}$ holds true while $\mathcal{R}_{test} \subseteq \mathcal{R}_{train}$.
    \item[\textbf{Fully-inductive setting}] For every triple $(h,r,t) \in \mathcal{T}_{test}$, both $h \notin \mathcal{E}_{train}$ and $t \notin \mathcal{E}_{train}$ holds true while $\mathcal{R}_{test} = \mathcal{R'} \cup \mathcal{R}$ such that $\mathcal{R'} \cap \mathcal{R}_{train} = \emptyset$, $\mathcal{R} \subseteq \mathcal{R}_{train}$. They further differentiate the following two sub-settings: (i) fully-inductive with \emph{semi unseen relations}: predict $(h,r,t)$; $r \in \mathcal{R}_{test}$, and (ii) fully-inductive with \emph{fully unseen relations}: predict $(h,r,t)$; $r \in \mathcal{R'}$.
\end{description}

Their partially inductive setting is equivalent to the fully equivalent definition by Galkin \textit{et al.}, and their fully-inductive setting is a restricted variant of the truly equivalent definition by Gesese \textit{et al.} In this case, the terminologies are not only diverging, but directly contradicting one another.
 
Additionally, Geng \textit{et al.} further subdivide this setting into \emph{testing with semi unseen relations} and \emph{testing with fully unseen relations}. In particular, \emph{testing with semi unseen relations} boils down to the truly-inductive LP setting as defined by Gesese \textit{et al.}

Diverging definitions lead to two main caveats. As shown above, some naming conventions are shared between authors, while they actually refer to different experimental settings. 
Likewise, naming conventions may differ between authors, but they can refer to the same concrete experimental setting.

These diverging definitions make it hard to figure out what the experimental, ILP-related setting actually is. It also makes performance comparisons more difficult. One actually has to carefully read the authors' definitions before knowing what the task at hand is. This is highly undesirable. We would like to have a clear terminological set where each task unambiguously maps to one concrete experimental setting.

\subsection{Towards a better understanding of knowledge-scarce link prediction}\label{sec:towards}
Before presenting our proposal for a novel nomenclature for non-transductive link prediction, we want to revisit the common distinction of ZSLP, FSLP, and ILP.
The necessity of clearly defining these tasks stems from our analysis in Section~\ref{sec:systematic}, where we mentioned a few works referring to two of these tasks.

\sstitle{Zero-Shot Link Prediction} is by far the least prone to confusion. While there is an overall agreement that ZSLP is characterized by the absence of any support set that would reveal connections between emerging elements and those seen during training, there is an implicit bias towards considering ZSLP as focusing on unseen \emph{relations}. As ZSLP originates from ZSL -- which was firstly defined in a very general framework of unseen labels appearing during testing~\cite{larochelle2008,palatucci2009,lampert2009} -- one might argue that there is no reason to limit ZSLP to unseen relations. In theory, ZSLP could concern unseen entities, relations, or both, provided that no training data is available or used to learn their representations.

\sstitle{Few-Shot Link Prediction} can be seen as a middle ground between ZSLP and the purely transductive setting. Like ZSLP, FSLP focuses on unseen relations, and testing on these relations usually follows an ordering into different episodes. 
However, compared to the ZSLP setting, FSLP is performed with the help of a limited and usually fixed number of training samples for each unseen element (\textit{i.e.} the few-shot examples). This means that shallow representations are actually learnt for ``unseen'' elements based on the support set, which raises the question of how this delineates from transductive setups (in particular such with a high relation imbalance and a subset of rare relations in the training set). 
Thus, FSLP could be better characterized as performing LP in a context of scarce information on a subset of entities and relations.

\sstitle{Inductive Link Prediction}, as discussed in the last section, covers many distinct settings. Overall, inductive models aim to embed entities and relations from the training graph into a low-dimensional continuous vector space and learn an inductive embedding generator, which can generate embeddings for unobserved, emerging entities and relations at testing time. This embedding generator is not designed to reason from specific training examples to specific test examples. In contrast, it is \emph{inductive} because the patterns learnt on training samples should lead to \emph{general} rules or representations to be later applied to the test examples, which is in line with how induction is defined.

\sstitle{Inductive vs. Few-Shot and Zero-Shot Link Prediction.} 
As mentioned in Section~\ref{sec:systematic}, a few works identify themselves as both FSLP and ILP. 
In light of our thoughts on FSLP above, this deserves questioning. 
Indeed, if such models have few-shot examples and are trained on them, the predictive task is generally not inductive. 
In contrast, the absence of examples in ZSLP corresponds to an inductive task.
Hence, FSLP, ZSLP, and ILP may not be disjoint: FSLP only denotes the existence of support triples on which training can be performed, ZSLP denotes their absence, and ILP is concerned with the model ability to learn rules or representations that are general enough to perform predictions on unseen entities and relations. In other words, FSLP and ZSLP denote the relative paucity of training data for emerging entities and relations and are thus concerned with \emph{training}. ILP, in contrast, is concerned with \emph{testing}. Accordingly, it has been said that many inductive models actually require a support set, \textit{i.e.} edges linking unseen entities appearing at testing time with seen entities from the training graph~\cite{markowitz2022ilp}. This means that an inductive LP model could well be few-shot or zero-shot, based on whether such support triples are available. However, due to the apparent implicit disjointness between ILP, FSLP, and ZSLP\footnote{As only 4 papers out of the 129 selected papers identify themselves as belonging to two different settings.}, a reasonable desideratum could also be to keep these settings unique and have their experimental conditions fully defined. 
We leave this for future work but suggest to rely on some distinctions, \textit{e.g.}, between ILP and FSLP: while under FSLP the support triples are used for learning shallow representations (meta-training), under the ILP settings these support triples -- if any -- are only used for revealing the connections between in-KG and out-KG elements at test time\footnote{For GNN-based models, this is often necessary for initializing their representations with embeddings of the k-hop neighbors.} but are not used for training purposes.

\subsection{An anchor-based nomenclature for non-transductive link prediction}

In light of the previous discussions and the issues that arise from using differentiated terminologies to define ILP, FSLP, and ZSLP, our purpose is to propose a concise yet comprehensive naming convention to refer to such settings, so that any ambiguity is resolved, and any overlap between settings is covered by different designations.

\sstitle{Approach.}
We claim that the nature of the unseen elements in testing triples could fully define the setting at hand. This fine-grained perspective based on the nature of test triples ensures that the model is tested on a collection of triples that comply with the exact same pattern of seen and unseen elements, which seems necessary to avoid any sources of entanglement between settings. For a given test triple $(h,r,t)$, some of $h$, $r$, or $t$ may have been seen during training. We call such elements \emph{anchors}, as they supposedly help making predictions on triples that contain unseen elements.

\begin{figure}[t]
  \centering
  \includegraphics[width=0.5\textwidth]{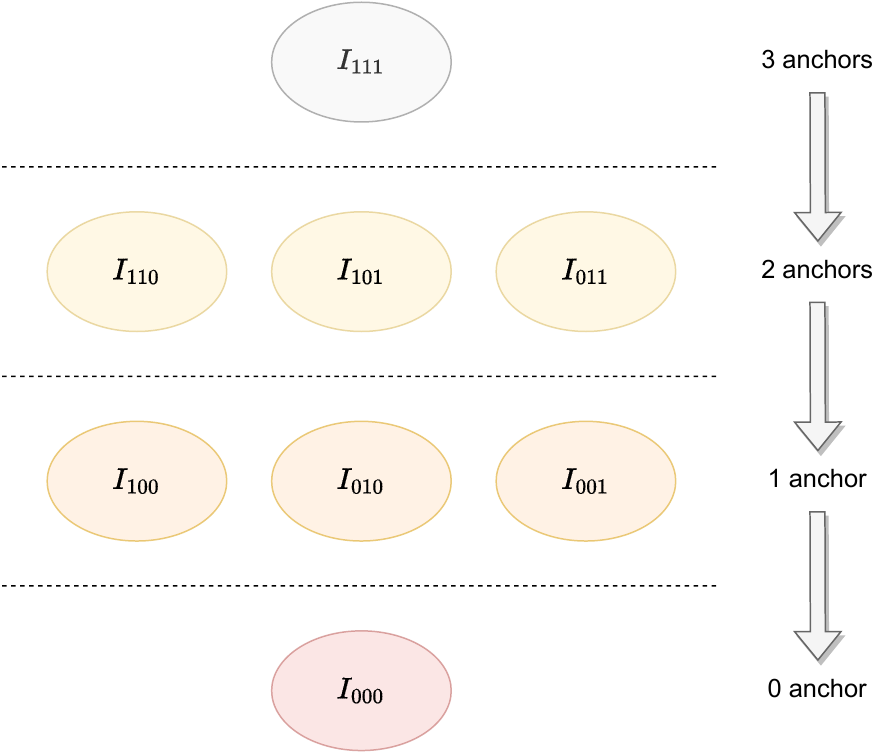}
  \caption{Our proposed anchor-based nomenclature}
  \label{fig:layers}
\end{figure}

Figure~\ref{fig:layers} depicts the general idea and the possible combinations of seen and unseen elements in test triples. More specifically, the indices 0 and 1 are chosen to materialize unseen and seen elements, respectively. The position of indices follows the natural ordering of triples' elements. As an example, the indices in the setting $I_{101}$ in Table~\ref{tab:alignment} refer to the head (1:seen), relation (0:unseen), and tail (1:seen), respectively. It is worth mentioning that this nomenclature is similar to what Oh \textit{et al.}~\cite{oh2022} briefly described in their work. However, they do not discuss the possibility of using such notation to consistently and systematically describe existing settings.

In Table~\ref{tab:alignment}, our proposed boolean indexing-based nomenclature is used as a structured reading grid to better compare settings from~\cite{galkin2022, daza2021ilp, geng2023, gesese2022, zhang2023DEKG} on the basis of test triples' \emph{signature}. For instance, test triples that are composed of seen heads, seen relations, and seen tails ($I_{111}$) \emph{may} appear in the semi-inductive LP setting as defined by Gesese \textit{et al.}, but are not considered in the other four works.
It should be noted that some settings as they are currently defined, cannot be fully described by one single item of our nomenclature. In such a situation, an ensemblist approach is adopted to comprehensively describe the complete range of possibilities. As an example, the semi-inductive LP setting from Galkin~\textit{et al.} is mapped to the $I_{011} \cup I_{110} \cup I_{010}$ setting. This is made possible as our 8 settings do not overlap regarding which sort of triples forms the testing set.

\begin{table}[h]
  \centering
  \caption{Comparing existing works under the lens of our proposed anchor-based nomenclature}\label{tab:alignment}
  \centering
  \footnotesize
    \resizebox{1.0\textwidth}{!}{
    \begin{tabular}{llllll}
    \hline
    \textbf{} & \textbf{Galkin \textit{et al.}~\cite{galkin2022}} & 
    \textbf{Daza \textit{et al.}~\cite{daza2021ilp}} & \textbf{Gesese \textit{et al.}~\cite{gesese2022}} & \textbf{Zhang \textit{et al.}~\cite{zhang2023DEKG}} & \textbf{Geng \textit{et al.}~\cite{geng2023}} \\
    \hline
    ${I}_{111}$ & \dashed & \dashed  & semi-inductive & \dashed & \dashed \\
    ${I}_{011}$ & semi-inductive & dynamic & semi/truly-inductive & bridging/extended & \dashed \\
    ${I}_{110}$ & semi-inductive & dynamic & semi/truly-inductive & bridging/extended & \dashed \\
    ${I}_{001}$ & \dashed & \dashed & truly-inductive & \dashed & \dashed \\
    ${I}_{100}$ & \dashed & \dashed & truly-inductive & \dashed & \dashed\\
    ${I}_{010}$ & semi/fully-inductive & dynamic/transfer & fully/truly-inductive & enclosing/extended & partially/fully\footnote{Subsetting semi-unseen relations.} inductive \\
    ${I}_{000}$ & \dashed & \dashed & truly-inductive & \dashed & fully\footnote{Subsettings semi-unseen relations and fully unseen relations.} inductive \\
    ${I}_{101}$ & \dashed & \dashed & \dashed & \dashed & \dashed \\
    \hline
  \end{tabular}
  }
\end{table}

\sstitle{Consistent and Systematic Mapping of Existing Works.}\label{sec:nomenclature}
By revisiting the discussed works in ILP, FSLP, and ZSLP, and mapping these works to the proposed nomenclature depicted in Fig.~\ref{fig:layers}, we aim at answering the following research questions (RQs):

\begin{itemize}
    \item \textbf{RQ1.} How terminology differs between recent works? In particular, are some areas more prone to terminological discrepancies?
    \item \textbf{RQ2.} Does aligning existing research with the proposed nomenclature reveals currently under-explored settings?
\end{itemize}

Results of this mapping are reported in Table~\ref{tab:mapping} an are further discussed in the following.

\begin{table}[t]
  \centering
  \caption{Systematic mapping of surveyed works to our anchor-based nomenclature. These works are categorized as FSLP, ZSLP, and ILP, according to how authors identify their works. Subsequently, these works are mapped to our anchor-based view w.r.t. the nature of test triples}\label{tab:mapping}
  \begin{tabular}{llll}
    \hline
    \textbf{Setting} & \textbf{FSLP} & \textbf{ZSLP} & \textbf{ILP}\\
    \hline
    $I_{111}$ & $53$ & \dashed & \dashed\\
    $I_{101}$ & \dashed & $13$ & $2$ \\
    $I_{011}$  & \dashed & \dashed & $3$ \\
    $I_{010}$  & \dashed & \dashed & $34$ \\
    \hline
    $I_{110} \cup I_{011}$  &\dashed & \dashed & $5$ \\
    $I_{010} \cup I_{000}$ & \dashed & \dashed & $1$ \\
    \hline
    $I_{110} \cup I_{011} \cup I_{010}$  & \dashed & \dashed & $11$ \\
    $I_{111} \cup I_{011} \cup I_{110}$ & \dashed & \dashed & $1$ \\
    $I_{100} \cup I_{001} \cup I_{000}$  & \dashed & \dashed & $2$ \\
    $I_{110} \cup I_{100} \cup I_{101}$ & \dashed & \dashed & $1$ \\
    $I_{011} \cup I_{001} \cup I_{101}$ & \dashed & \dashed & $1$ \\
    \hline
\end{tabular}
\end{table}

\sstitle{RQ1.} 
As stated above, one characteristic of FSLP can arguably be that some training is performed on support triples (during meta-training) to learn representations for emerging elements. Thus, in our nomenclature, FSLP is closer to the transductive setting than it is to ZSLP and ILP, due to the fact that ${I}_{111}$ systematically maps to all the retrieved FSLP works. Consequently, the way authors define FSLP and identify themselves to this setting remains consistent with our mapping convention. Similarly, all surveyed ZSLP-related works map to the ${I}_{101}$ anchor-based setting. Therefore, FSLP and ZSLP seem to have consistent conceptualizations in existing works, as mapping these works to our nomenclature does not reveal any discrepancy. In contrast, ILP-related works span across a broad range of experimental configurations w.r.t. the nature of test triples. Most of the works map to $I_{010}$, but many of them also map to $I_{110} \cup I_{011} \cup I_{010}$ and, secondarily, to $I_{110} \cup I_{011}$. This is highly detrimental to understanding what the evaluation procedure is. If all these works identify themselves as performing ILP, any model comparison becomes spurious and irrelevant. Our proposed anchor-based nomenclature has the benefit of clearly delineating these works from each other.

\sstitle{RQ2.} Table~\ref{tab:mapping} provides a quick overview of over-represented anchor-based settings. In terms of settings featuring a single type of test triples, we observe that over the 6 possible combinations, only $I_{111}$, $I_{101}$, $I_{011}$, and $I_{010}$ are actually represented in existing works. It should be noted, however, that $I_{110}$ can be seen as the counterpart of $I_{011}$ if relations are reversed. Importantly, Table~\ref{tab:mapping} reveals that the $I_{000}$ setting has not been explored in surveyed papers. To the best of our knowledge, this setting could have never been explored altogether. Having test triples featuring only unseen elements has already been deemed as a challenging task~\cite{geng2023}. Such test triples are studied in the $I_{010} \cup I_{000}$ and $I_{100} \cup I_{001} \cup I_{000}$ anchor-based settings. However, (i) only three works actually consider them in settings mixing various types of test triples, and (ii) the $I_{000}$ setting is never studied independently. Regarding (ii), one reason could be the relatively low likelihood of that setting appearing in real-world applications: an emerging batch of triples featuring only unseen entities and relations is rather unlikely to happen and difficult to handle. Regarding (i), however, designing models able to handle both unseen entities and relations in at least some test triples is an under-explored research avenue that certainly deserves greater consideration. 

\section{Conclusion}\label{sec:conclusion}
In this survey, we thoroughly reviewed papers related to ILP, FSLP, and ZSLP.
Concerned with binary relational facts, this survey analyzed the main trends w.r.t. the ILP, FSLP, and ZSLP tasks. This led us to discuss crucial considerations around benchmark datasets, task definitions and terminology. Ultimately, a pragmatic nomenclature is presented and used to consistently map each work to a unique, unambiguous name. This survey intends to pave the way for further analysis and discussions related to these settings. Both their recent emergence and growing significance in contemporary works encourage to better frame the field of knowledge-scarce LP as a whole and provide a comprehensive and unified understanding of the three aforementioned settings.

\bibliographystyle{ACM-Reference-Format}
\bibliography{sample-base}

\appendix

\end{document}